# Knowledge-Integrated Informed AI for National Security

*Anu K. Myne*
*Kevin J. Leahy*
*Ryan J. Soklaski*

This page intentionally left blank.

# ABSTRACT

The state of artificial intelligence technology has a rich history that dates back decades and includes two fall-outs before the explosive resurgence of today, which is credited largely to data-driven techniques. While AI technology has and continues to become increasingly mainstream with impact across domains and industries, it's not without several drawbacks, weaknesses, and potential to cause undesired effects. AI techniques are numerous with many approaches and variants, but they can be classified simply based on the degree of knowledge they capture and how much data they require; two broad categories emerge as prominent across AI to date: (1) techniques that are primarily, and often solely, data-driven while leveraging little to no knowledge and (2) techniques that primarily leverage knowledge and depend less on data. Now, a third category is starting to emerge that leverages both data and knowledge, that some refer to as "informed AI." This third category can be a game changer within the national security domain where there is ample scientific and domain-specific knowledge that stands ready to be leveraged, and where purely data-driven AI can lead to serious unwanted consequences.

This report shares findings from a thorough exploration of AI approaches that exploit data as well as principled and/or practical knowledge, which we refer to as "knowledge-integrated informed AI." Specifically, we review illuminating examples of knowledge integrated in deep learning and reinforcement learning pipelines, taking note of the performance gains they provide. We also discuss an apparent trade space across variants of knowledge-integrated informed AI, along with observed and prominent issues that suggest worthwhile future research directions. Most importantly, this report suggests how the advantages of knowledge-integrated informed AI stand to benefit the national security domain.

This page intentionally left blank.

# ACKNOWLEDGMENTS

I'd like to thank Robert Bond and Sanjeev Mohindra for their enduring support and encouragement. Special thanks also goes to my co-authors, Kevin Leahy and Ryan Soklaski, for their practical viewpoints and for supporting the extensive literature review, and to all cited authors who are making great contributions to the advancement of artificial intelligence technology.

This page intentionally left blank.

# TABLE OF CONTENTS

This page intentionally left blank.

# LIST OF FIGURES

# LIST OF FIGURES
## (Continued)

# LIST OF TABLES

This page intentionally left blank.

# 1. MOTIVATION AND SCOPE

Artificial intelligence (AI) techniques are numerous with many unique variants that employ different learning algorithms and architectures. One can separate techniques by the two resources that these techniques predominantly leverage: (1) techniques that focus on leveraging knowledge with less or no dependence on data, or (2) techniques that are primarily, and often solely, data-driven. The main diagonal of the matrix in Figure 1 captures these two overarching categories in the top left and bottom right. Historically, the former set of approaches that directly capture knowledge (scientific, mechanistic, or domain expertise) were dominant. Recently though, data-driven approaches that learn patterns from data have taken center stage for many applications, especially speech recognition [3], game play [4], and image classification [5].

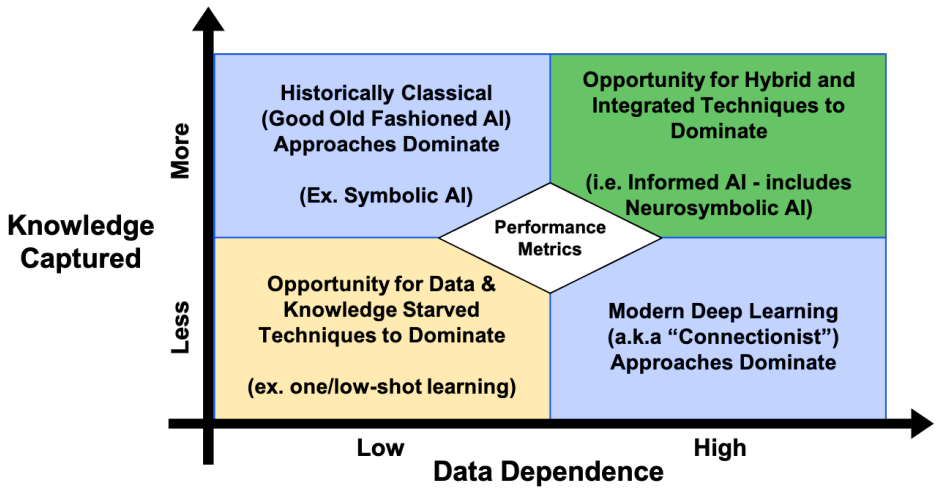

**Figure 1.** *Grouping AI Techniques by Knowledge Captured and Data Dependence*
*This 2x2 matrix divides the space of AI techniques by the degree of knowledge captured and the level of data-dependence. Regardless of class of technique, getting and staying ahead with AI requires performance across several metrics including accuracy, explainability, and reliability, to name a few. Knowledge-Integrated informed AI techniques, which fall in the top right (green) quadrant, promise noteworthy gains across several key performance metrics. Foundation models [6], an almost exactly opposing paradigm, fall in the lower right corner of the bottom right quadrant.*

While numerous promising research directions exist in both major categories, a third class of techniques that are hybrid (exploiting both knowledge and data in fundamentally new ways) is emerging with promise to achieve profoundly more capable AI (top right quadrant of matrix). We refer to this broad space, which encapsulates it's own set of sub-classes of techniques, as "knowledge-informed AI." There is a long list of names given to these techniques, some listed to the right of Figure 2, which suggests that terminology in this paradigm of AI is unsettled. In their excellent survey paper published in 2021 [1], Laura von Rueden et. al. coined the term "informed machine learning," while others use terms like "scientific machine learning," "neurosymbolic AI," or "physics-informed AI." As illustrated in the Venn diagram in Figure 2, symbolic AI can be considered as a class of techniques that are inherently informed by knowledge, but that's not new nor is it part of the emerging paradigm we aim to shed light on in this report. Instead, if we take standard forms of deep learning and reinforcement learning and marry them with either symbolic AI or other non-AI models with embedded knowledge, it's the areas of intersection that make up the new and emerging paradigm that is the scope of this report. We focus very little on neurosymbolic hybrid AI, where the two worlds of data-driven and knowledge-based models are usually only loosely-coupled, and we focus heavily on techniques that we refer to as "knowledge-integrated informed AI."

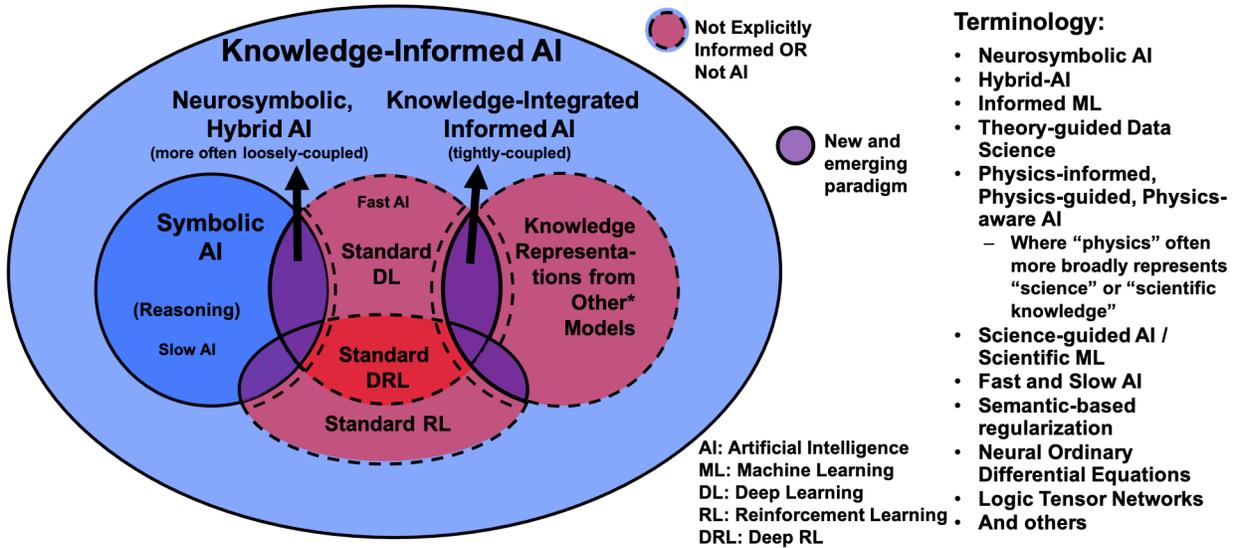

*Figure 2.* Venn Diagram of Informed AI
*Informed AI is a superset of approaches that includes symbolic AI, Neuro-symbolic AI, and Knowledge-Integrated Informed AI. The areas of intersection (purple) represent new and emerging AI paradigms. The focus of this report is on knowledge-integrated informed AI.*

Our definition of knowledge-integrated informed AI is the following:

**Knowledge-Integrated Informed AI**: AI techniques in which explicit, principled and/or practical knowledge that is scientific, domain-specific, or data-centric is integrated (not necessarily in a separable way) into a model's development pipeline. In these techniques:

- Knowledge from scientific or domain expertise is integrated to provide partial explanations or constraints for the unknown relationship that the model is built to learn,

  Or

- Understanding of properties of training data are exploited to provide more efficient/effective mechanisms for learning.

This definition is illustrated with the block diagram in Figure 3. While it can be standard practice to consider knowledge about the problem while developing machine learning models, the actual integration of explicit knowledge so that it constrains or guides the learning process is the major distinguishing factor. There are several forms of knowledge representation as well as paths of knowledge integration that are discussed later in this report.

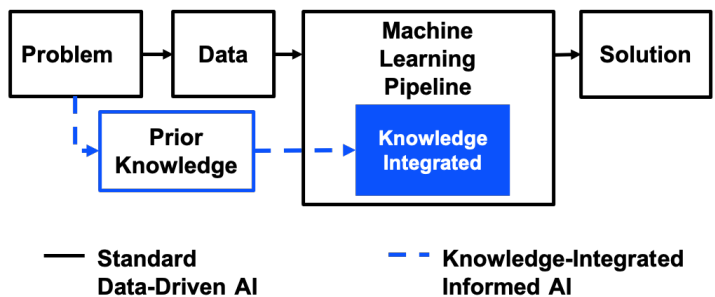

**Figure 3.** *Knowledge-Integrated Informed AI Block Diagram*
*The integration of explicit knowledge in data-driven AI pipelines is what distinguishes knowledge-integrated informed AI from standard data-driven AI.*

In some ways, the promise of knowledge-integrated informed AI techniques has already been proven. Figure 4 illustrates two well known, breakthrough developments that are precursors of this new paradigm. While the convolutional neural network changed the world of image recognition by exploiting data-centric knowledge (that low-level image features are local and exhibit translational invariance), AlphaGo changed the world of reinforcement learning by leveraging practical knowledge of the game being played (yes, the more recent MuZero masters game play without rules, but with far more compute power and time that can be considered unnecessary and wasteful). Both of these precursor examples that integrated different forms of knowledge through distinct entry points demonstrated superior performance and broad-reaching impact, which suggests that more deliberate integration of knowledge can lead to many more breakthroughs.

A more recent example that fits squarely into this new paradigm of AI is the Neural Circuit Policy model that has received media attention [7] [8] for its ability to "learn cause and effect" for a navigation task (Figure 5). The model learns to steer an autonomous vehicle with very few neurons, "superior generalizability, interpretability, and robustness compared to orders of magnitude larger black-box learning systems." Upon deeper investigation, we realize that this model has knowledge-integrated design elements that play a big hand in its ability to achieve noteworthy performance. It has a bio-inspired architecture where the final output is arrived at in steps that go from sensor to decision to actuation, and the network is followed with an ordinary differential equation (ODE) solver that captures the dynamics of the nervous system to "obtain numerically accurate and stable solutions." Without these design elements, the performance of this model would undoubtedly be unremarkable. [9]

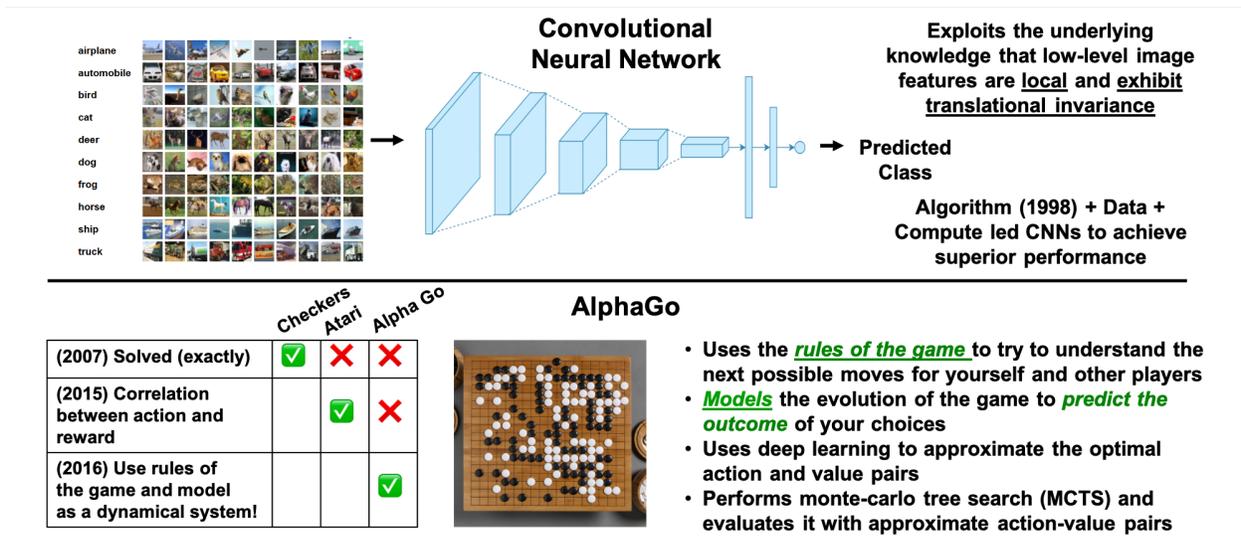

*Figure 4.* Well-known, Early Examples of Knowledge-Integrated Informed AI
Both the convolutional neural network and AlphaGo are well-known examples of breakthroughs in deep learning and reinforcement learning that leverage knowledge about the problems they are designed to solve.

*Figure 5.* Recent Example of Knowledge-Integrated Informed AI
The Neural Circuit Policy model designed by Lechner et. al. is a nice recent example of groundbreaking performance achieved by a novel knowledge-integrated informed AI approach.

Knowledge-integrated informed AI techniques are particularly appealing for the national security domain, where there are elevated concerns with purely data-driven approaches that often lack in explainability, generalizability, traceablity, and reliability. Intuitively, techniques that exploit both data and knowledge, where knowledge can be leveraged to satisfy constraints or guide learning, *should* provide superior performance (such as improved accuracy, reliability, robustness, and more efficient use of resources). It turns out that many recent works, in addition to the examples discussed above, have now proven this intuition. Furthermore, while there exist opposing perspectives, several reputable scholars have suggested that knowledge-integrated informed AI will bring the next generation of AI, or its third wave [10] [11]. In subsequent sections in this report, we explore numerous recent examples of knowledge-integrated informed AI, what's behind these claims of game-changing potential, and what this new paradigm of AI can mean for problems of national security.

In order to thoroughly explore the current state of the art of knowledge-integrated informed AI for national security, the literature review summarized in this report considers many published examples. Our effort was hugely facilitated by the taxonomy presented in Laura von Rueden et. al.'s 2020 survey paper [1], as well as a few other recent and well formulated surveys [12] [13] [14]. Subsequent sections in this report dive into the various knowledge-integrated informed AI architectures, distinct paths of integrating knowledge into AI pipelines, illuminating examples and their reported performance gains, observed research gaps and opportunities, and how knowledge-integrated informed AI could positively impact the national security domain.

## 1.1 FURTHER DISCUSSION OF COMPUTER MODELS AND AI PARADIGMS

We referred earlier to two broad categories of AI. Here we consider the individual strengths and challenges of knowledge-based models and data-driven AI models, what's complementary and opposing, and how these arguments lead to a hybrid class.

First, the following are our definitions of two classes of computer models:

- We consider <u>knowledge-based models</u> to broadly include models that capture scientific or mechanistic phenomena, or human expertise, through specific representation of fundamental understandings of a domain. Generally, these models are designed to serve a multitude of purposes that need some representation of the world to perform their assigned tasks. In terms of AI models, symbolic AI (reasoning based on symbolic knowledge) and other classical AI models (ex., logic-based models and probabilistic graphs) fall into this category. As defined, this class of computer models is more broadly inclusive, and therefore also includes all kinds of models that capture and represent knowledge, even non-AI models.

- We define <u>data-driven AI models</u> as essentially universal function approximators that learn from data, such that model architectures and/or weights are updated via learning pipelines have no explicit consideration of domain knowledge. Generally, these models are designed to perform various tasks such as perception, reasoning, planning, and even to represent and understand the world. Learning models that also exploit explicit knowledge in addition to data do not fit into this category.

Many in the AI community refer to an earlier paradigm of AI being one of symbolism, or reasoning based on symbolic knowledge (a paradigm that dominated until the 1980s), and the more recent paradigm as one of connectionism, or data-driven decision making using neural networks (which began increasing in popularity in the 1990s). While connectionist models are exactly analogous to data-driven AI paradigm, referring to only symbolic models as the opposing knowledge-informed AI paradigm paints an incomplete picture, because knowledge can be represented in non-symbolic ways. But the fusion of symbolic and connectionist AI is gaining similar attention as the fusion of knowledge-models with data-driven AI models. So this discussion here draws comparisons between both symbolic and connectionist models as well as knowledge-informed and data-driven models.

All classes of computer models in discussion here have been and continue to be used extensively across domains, and have demonstrated their usefulness. With their respective strengths, they each come also with many challenges that deem them unfeasible or insufficient for certain applications.

There is undoubtedly decades of knowledge and domain expertise built into many high-performing knowledge-based models. To their advantage, these models are able to represent and reason about complex, causal, and abstract concepts with solutions that are traceable, interpretable and explainable. These valuable strengths however, come with design processes that are all-too-often manually intensive and highly problem-specific, with many opportunities to insert biases, and even preclude discovery of new insights. Furthermore, problems can more quickly become computationally unfeasible as cost to run models at appropriate resolutions in space and/or time is often high, and the number of samples needed to provide good estimates can be prohibitively expensive [12] [15] [16].

For example, in symbolic systems that are developed for classification problems, rules or other non-numerical methods are used to reason about properties (often abstract) of objects to draw conclusions about them. One common example is a decision tree. In the case of cat classification, one can imagine a decision tree that asks if an animal is a vertebrate, a mammal, etc., to determine whether it is a cat. A well-known example of such classification systems is taxonomy in biological systems, such as the Linnaean taxonomy. The strength of such symbolic models lies in their interpretability and explainability. They are very transparent, and they can reason about arbitrary and abstract properties. However, while these symbolic models have strong ability to reason, they are not good at processing data. For example, there is not a natural way for a decision tree to classify the contents of an image.

Similar to knowledge-based models, data-driven AI models have clearly showcased winning performance and impact. What formulates these models can be viewed as incredibly versatile—more general-purpose and less problem-specific yet still providing desirable performance. Their versatility is largely attributed to the pliable nature of the learning framework and powerful constituting elements. As much as they are appreciable, they also exhibit several drawbacks. Training the models requires massive datasets (or intensive interaction with the environment for reinforcement learning) which are easily entrenched with a variety of imperfections (noise, incomplete coverage, lack of variability/representation, etc.), and the final product often fails to generalize beyond its training distribution [17]. What machines learn via purely data-driven approaches is a direct product of the imperfect data combined with algorithms that cannot compensate for those imperfections. While learning may be layers deep, actual understanding is easily only surface level or just indemonstrable. Furthermore, like knowledge-based models, this class of AI also has many entry points for bias (data collection or conditioning, the learning algorithm, and even the network architecture), and comes with computational drawbacks, such as an alarming amount of energy needed to supply the staggering amounts of compute power used for model development [18] [19]. Furthermore, their lack of interpretability, and explainability, and their sensitivity to adversarial attack raise many more alarms.

To continue with classification as the example problem, rather than reasoning through the task, a connectionist model (a.k.a., a data-driven deep learning model) performs via numerical representation and operations that use lots of data to learn relationships between inputs and outputs. For example, to classify images of cats, a neural network typically needs to be shown a very large number of images, with labels indicating the presence or absence of cats. Generally, this need for a large amount of data is not present in symbolic AI. Once a network has the data, it performs numerical operations on the data. In the image classification example, the system performs matrix operations on pixel values in order to find commonalities among similarly labeled images. The resulting trained neural network can determine whether a previously unseen image does or does not contain a cat. It is good at perceiving and processing data, but its process for classifying is opaque.

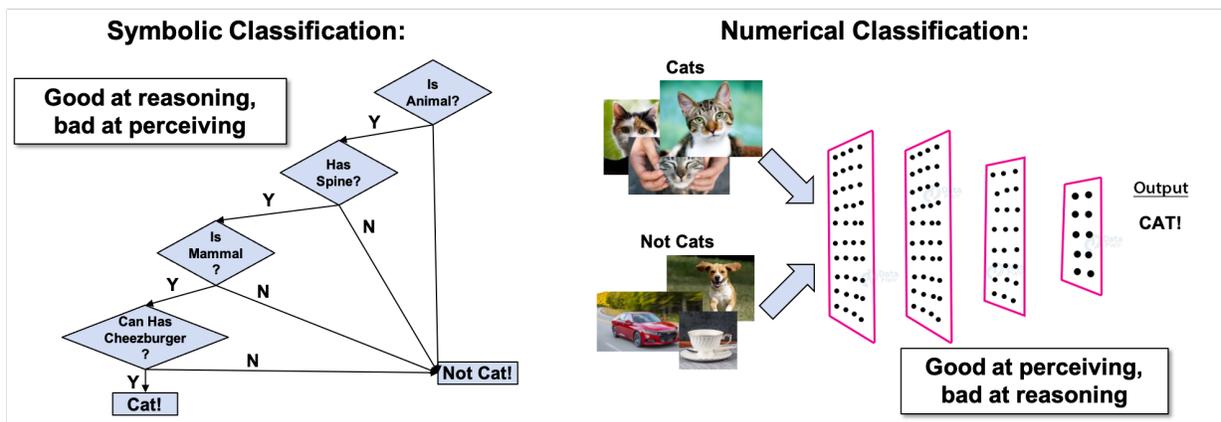

*Figure 6.* Symbolic vs. Numerical Classification
*A decision tree as an example of a symbolic cat classifier (left), and a neural network as a numerical cat classifier (right). Symbolic classification reasons directly on properties (possibly abstract) using methods based on "symbolic" (human-readable) representations of problems, logic, and search, where symbols are manipulated symbolically in otherwise purely numeric computation frameworks [10]. Conversely, in numerical classification, an image classifier learns to classify based on relationships between pixels in an image file using numerical operations.*

As the strengths and challenges of knowledge-based models and data-driven AI models lead to a combined class of knowledge-integrated informed AI, so do the strengths and challenges of symbolic and connectionist AI lead to neurosymbolic AI.

In the case of knowledge-integrated deep learning, these models are more accurate, generalizable, interpretable, data-efficient and computationally-efficient because of the knowledge infusion into otherwise black-box, unaware neural networks. Their learning is guided by things we know about the relationship that needs to be learned or, if not course correcting, then knowledge is constraining or restricting what the models learn tying them down with known aspects of the relationship that has unknowns to learn.

In the case of neurosymbolic AI, models possess reasoning abilities of symbolic systems and the perception and data processing abilities of numerical systems. In terms of the cat classification problem, consider the case of distinguishing among closely related species of rare wild cats. There may not be enough imagery of any one species to train a neural network. But there may be too much imagery (e.g., from cameras placed in remote locations to capture footage of wild animals) for a human to process alone. In that case, humans can use their expertise to build a decision tree about different traits that each species of cat has—e.g., tufted ears, stripes, long whiskers, etc. Imagery of different species can be used to train a neural network classifier on each of those traits. For a new image, the results of the classifiers can be processed on a decision tree to determine if the image contains one of the rare species of interest. Thus, the perceptive neural network feeds the decision tree, which can reason about the output. Figure 7 illustrates a neurosymbolic concept.

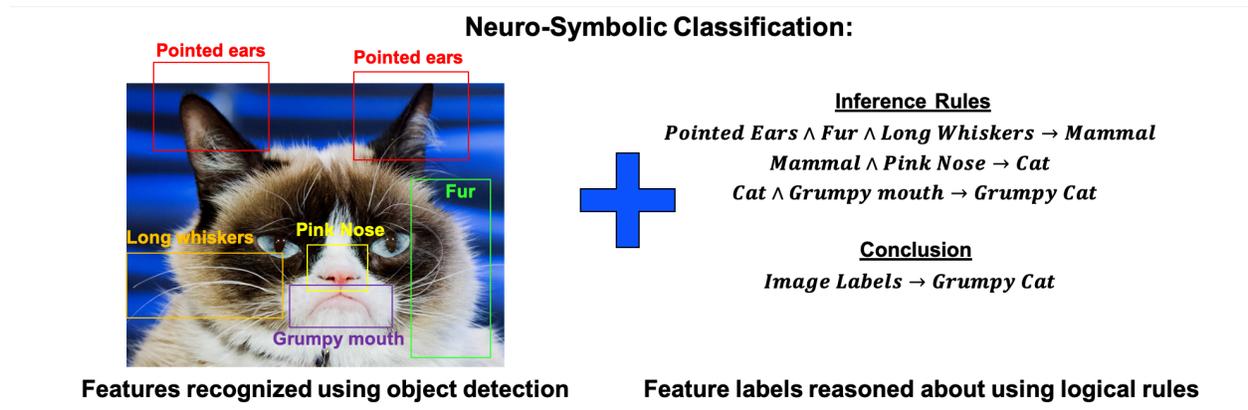

*Figure 7. Neurosymbolic Classification*
*Neurosymbolic AI possesses the reasoning abilities of symbolic systems and the perception and data processing abilities of numerical systems.*

## 1.2 AN OPPOSING AI PARADIGM

This discussion of emerging AI paradigms would be incomplete without reference to an entirely opposite research direction that aims for AI that does not leverage knowledge.

### 1.2.1 Foundation Models

Foundation models are also at the forefront of AI research, especially for natural language processing (NLP) and computer vision (CV) applications [6]. These models eschew the use of highly-regimented model architectures (e.g., recurrent layers for NLP applications, and convolution layers for CV) trained in a supervised setting, in favor of less-structured architectures that learn in a self-supervised manner. Models like BERT [20], GPT3 [21], and CLIP [22] are among the most widely recognizable and influential foundation models. These approaches demonstrate that the availability of large amounts of highly varied, unlabeled data—even if the data is very noisy—can be sufficient to produce a model that can adapt to novel tasks in a zero-shot manner and that achieves state-of-the-art performance on these tasks.

The most costly and obstructive aspect of supervised deep learning is the need for large, hand-labeled data sets to inform one's model. Foundation models eliminate this cost altogether, as they have proven capable of learning from large, multi-modal repositories of structured (but unlabeled) data (e.g., Flickr [23] and Wikipedia [24]). However, these public data repositories are, themselves, unrivaled in size: they are the culmination of decades-long contributions from millions of individuals from across the globe. Furthermore, the process of training these models represents a paradigm-shift toward being increasingly sophisticated and prohibitively expensive. For instance, a standard ResNet-50 CV model can be trained in a supervised fashion to achieve competitive performance on ImageNet using only 64 GPU hours [25]. By contrast, training CLIP to achieve state-of-the-art performance on ImageNet zero-shot required 86,000 GPU hours [26]. It must be emphasized that this is **not** an apples-to-apples comparison, and that the resulting CLIP model is fundamentally distinct in nature, as it is capable of performing far more generalized tasks than is the ResNet model. That being said, the 1,000x difference in computational cost to develop the model is resounding nonetheless.

It remains to be seen how foundation models will impact the role of deep learning in scientific fields, where knowledge-informed AI techniques can more naturally find purchase [27].

### 1.2.2 Reinforcement Learning without Knowledge

In recent years, some researchers—most notably DeepMind—have made an effort to reduce the knowledge used in deep reinforcement learning (Figure 8). Like foundation models, this approach is in direct opposition to the notion of knowledge-informed AI.

DeepMind developed AlphaGo in 2016 [4], which learned to play the game Go using data from human gameplay, domain knowledge, and the rules of the game. They have progressively pared back the knowledge provided to their system. Notably, AlphaZero [28] used a single algorithm for multiple games, although it had rules for all three games it learned. To date, this effort has culminated in MuZero [29], which learned four games with a single algorithm, without knowledge of the game. MuZero learns three models for playing a game (Figure 9): a representation of the state, the dynamics of the game, and the policy and reward. In a typical reinforcement learning scenario, the state and reward are chosen by a user or designer, and even the dynamics may be known. In that case, the system only needs to learn the policy. Here, the system learns the three components from scratch.

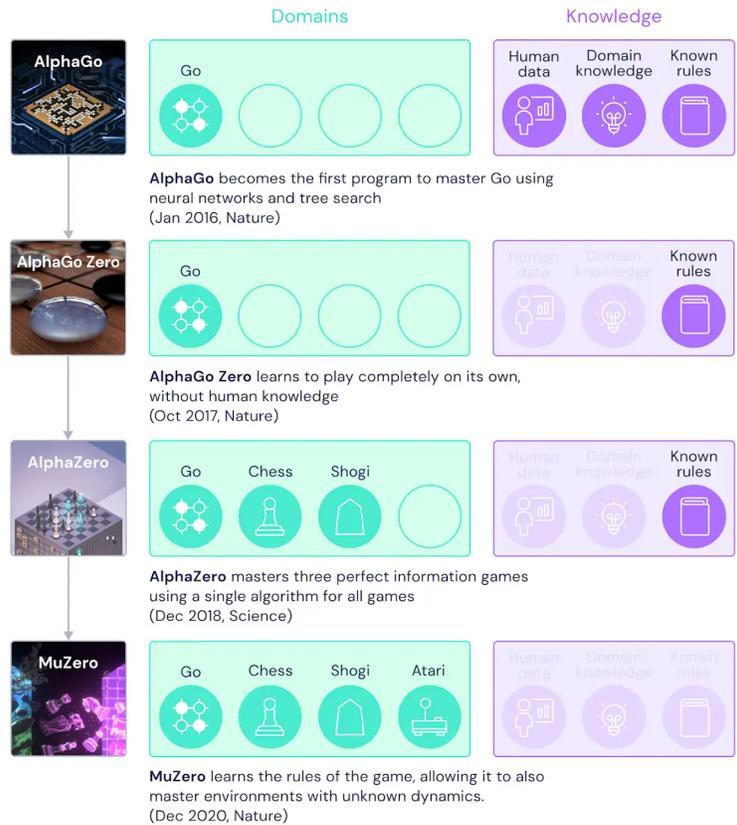

***Figure 8.*** *Removing Knowledge from the RL Pipeline*
*Since 2016, DeepMind has pursued reinforcement learning that uses increasingly less knowledge. This line of research, starting with AlphaGo and culminating in MuZero in 2020, provides an approach for mastering several games with no human data, domain knowledge, or rules. This approach is in direct contrast with an Informed AI approach. [30]*

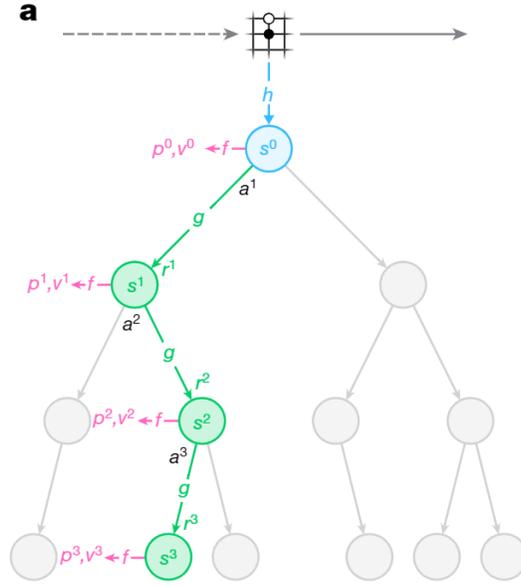

**Figure 9.** *Reinforcement Learning without Prior Knowledge*
*MuZero learns three separate models for playing games. Function h is an encoding of the current state, g models the dynamics of the game, and f models the policy and reward. [29]*

We note that MuZero uses a single algorithm to learn a variety of games without any prior knowledge. Unlike foundation models (see Section 1.2), it does not generalize from one game to another. Foundation models provide a *trained model* that can be modified or tuned to suit a variety of problems. MuZero, on the other hand, provides an *algorithm* that is broadly useful for many games, but needs to be separately trained on each game. We note also that the computational results reported for MuZero require many hours or days of specialized computational hardware. One goal of informed AI is to reduce computational requirements by informing the system.

This page intentionally left blank.

# 2. KNOWLEDGE-INTEGRATED INFORMED AI FOR NATIONAL SECURITY

As research and development of AI solutions has become increasingly prevalent toward addressing problems of national importance, the performance requirements of these systems have been evolving. We have realized that we need AI that not only performs accurately, but that is also traceable (i.e., correct-by-construction, and interpretable), reliable (i.e., generalizable to novel instances), and robust against adversarial attack [31]. Some of these metrics are often posed as ethical guidelines, but they are also markers for well-engineered systems that enable better human-machine teaming. **Bottom line up front: Utilizing valuable knowledge from the extensive range of scientific and domain expertise that exists within the national security domain in knowledge-integrated informed AI solutions will help realize many AI system performance goals.**

Technology for national security is a broad mission that includes many diverse sub-domains (ex., air/land/sea/space defense, homeland protection, air traffic control, cyber security, intelligence, surveillance, reconnaissance, defense against biological and chemical weapons, etc.). A vast and constantly evolving set of technologies are developed to address important problems associated with each of these mission areas, which all leverage one or more of the following knowledge-bases:

- Scientific Expertise (ex., physics, biology, chemistry, etc.)
- Scientific/Expert Understanding of Technologies (how they are designed, fabricated, their limits, etc.)
- Expert Understanding of the Applications for which Technologies are Developed

Nearly all formal representations of knowledge (in the form of equations, dynamical systems, probabilistic relations, logic rules, knowledge graphs, simulations invariances, or human feedback), which are presented later in this report (Section 3.2), come into play for each of these major categories of knowledge in the technology for national security domain. As AI is making headway across this domain, leveraging these knowledge sources more deliberately and explicitly can be game changing. Table 1 suggests just a few example ways these knowledge sources may integrate with AI and prove to be strategically advantageous.

## 2.1 PERFORMANCE GAINS REALIZED VIA KNOWLEDGE-INTEGRATED INFORMED AI

The hypothesis going into this study was that integration of knowledge into typically data-driven AI should yield significant gains for specific performance metrics that matter for problems of national security—at a minimum, producing more trustworthy AI because it satisfies, is consistent with, or is guided by knowledge defined by science or domain experts. The idea was that this kind of knowledge integration can lead us to more capable AI that requires less data, is more explainable with better interpretability, and generalizes better to novel samples unseen during training.

# TABLE 1

## Opportunities in Terms of Knowledge Sources

| **Knowledge Sources and Likely Representations:** | | Possible Knowledge-Integrated Informed AI Goals / Use Cases |
|---|---|---|
| Scientific Expertise (ex. Physics, biology, chemistry): | | - Use domain knowledge (simulators, governing equations, dynamical systems, relations, associations, etc.) to ground models that aim to discover/learn from data |
| Algebraic Equations | Knowledge Greaphs | |
| Differential Equations | Simulations | |
| Probabilistic Equations | Invariances | |
| Logic Rules | Human Feedback | |
| Scientific/Expert Understanding of Technology: | | - Develop neural network surrogate solutions that demonstrate consistent/accurate scientific phenomena with computational speedup and/or higher-order approximations |
| Algebraic Equations | Knowledge Graphs | |
| Dynamical Systems | Simulations | |
| Probabilistic Equations | Invariances | |
| Logic Rules | Human Feedback | |
| Expert Understanding of Technology Applications: | | |
| Algebraic Equations | Knowledge Graphs | - Employ AI solutions that are constrained/guided by policy/rules of engagement |
| Discrete-time Models | Simulations | |
| Probabilistic Equations | Invariances | |
| Logic Rules | Human Feedback | |

TABLE 2

Performance Gains Achieved by Knowledge-Integrated Informed AI

| Performance Gain | Quantified / Qualified Gain |
|---|---|
| **Greater Accuracy** | +0.6-20% Accuracy, -(4-70)% Test Error, 1-8x Score |
| **Extended Applicability** | Implied |
| **Enabling New Capabilities and Discoveries** | Implied |
| **Improved Safety and Reliability:** | |
| - Correct-by-Construction | Implied |
| - Interpretable*, Traceable*, Explainable* | Claimed Qualitatively |
| - Robustness via Generalizability | Claimed Qualitatively (usually) |
| **More Efficient Use of Resources:** | |
| - Data Efficiency | 40-96% less data, 0-1% of labels |
| - Computational Speedup | 2-15,000x |
| - Network Size | 1/3rd parameters |

Exploration of the state-of-the-art, from examples that date back decades to now, confirmed these hypotheses along with additional unanticipated benefits. Knowledge-integrated informed AI, all variants included, is growing in popularity and has been shown to offer one or more of the gains in performance summarized in Table 2. Note that some improvements are only claimed qualitatively, while others vary and fall inside the spread that we aggregated from the subset of papers we reviewed.

The reasoning behind knowledge integration leading to an applicability extension beyond the already broad application space is twofold. Firstly, this can be attributed to the increases in accuracy and the mechanism to ground a model by the truths of a domain which is, in turn, leading to more domain experts adopting modern AI techniques (especially in scientific domains). Secondly, these approaches are opening up the application space of deep learning from mostly perception and classification or generative tasks, to neural network surrogates that can replace computationally expensive knowledge-based models (expensive physics-based or earth-science models, for example).

There were some expected performance gains that were not found in the literature. First, neither robustness to adversarial attack, nor deeper and/or contextual understanding came up in any of the reviewed papers. While these possible benefits have not yet been highlighted, they likely will be in the future. Furthermore, as we surveyed the literature, we postulated other additional benefits that also have not been found to be prevalent in the literature, but likely will be in the future. First, some explicit knowledge integration approaches may be more easily updated *new* to accommodate situational constraints to enabling better adaptability. Second, easier verification using more standard methods could be achieved for models with fewer neurons. Third, knowledge integration could improve a model's memory efficiency, i.e., ability to maintain fundamental "understanding" even if trained to perform multiple tasks (that generally can cause memory loss—a.k.a., catastrophic forgetting). Section 6 lists all of these performance gains that we found to be uncharted in the literature as worthwhile future research directions.

## 2.2 BENEFITING NATIONAL SECURITY APPLICATIONS

Many example national security applications stand to benefit from one or more of the performance gains offered by knowledge-integrated informed AI approaches. Table 3 lists several example applications while Table 4 offers some specific, but rudimentary, example problems.

The first row block in Table 3 suggests that we can expect greater accuracy with knowledge-informed AI for AI tasks that fall in or around a processing chain with an available data stream *and* associated valuable knowledge (formally representable). For example, many signal processing applications fit this description, along with traditional, computationally-expensive models that can entirely or partially be replaced with neural network surrogates. Furthermore, given this strong potential for improvements in accuracy, applicability to problems for which modern AI techniques have been previously rejected may now be reconsidered. Moreover, this broadening of applicability is leading to new capabilities and new discoveries.

The second row block in Table 3 suggests that improved safety and reliability, which come through techniques that are correct-by-construction, interpretable (through traceability and explainability), and more robust because of generalizability, will also be beneficial for many national security applications. Obviously benefiting applications are ones that are described as high-consequence and safety-critical, and those that rely on humans and machines performing as a team. Additionally, problems that present dynamic and unpredictable settings will benefit from greater generalizability, as will problems where the knowledge may be changing with time (explicitly integrated knowledge is more easily updated). Perhaps a less obvious benefiting application is of problems that need multi-task solutions, which can suffer from catastrophic forgetting, but can benefit from integration of strong and unchanging foundational knowledge.

Finally, the third row block in Table 3 suggests that many national security applications also stand to benefit from more efficient use of resources, whether the resource is data, compute power, or the size or simplicity of the neural network. These applications include those that belong to data-limited domains (many if not most, most if not all), and problems that use AI at the tactical edge, especially if the AI is deployed on a platform with restricting size, weight, and power (SWaP).

TABLE 3

**Opportunities in Terms of Performance Gains**

| Performance Gain Achieved | Benefiting National Security Domains/Applications |
|---|---|
| **Greater Accuracy** (+0.6-20% Accuracy, -(4-70)% Test Error, 1-8x Score)<br>**Extended Applicability** (Implied)<br>**Enabling New Capabilities and Discoveries** (Implied) | - Processing chains with available data streams and formally expressible knowledge: ex. Radar, SAR, GMTI, IR, SIGINT, LIDAR, GPS/GPS-denied, etc.; Materials, weather, climate, chemistry, biology, physiology, etc.; C2, IC, LE, HADR, IO, etc.<br>- Traditional, computationally-expensive models – NN surrogates |
| **Improved Safety and Reliability:**<br>- Correct-by-Construction (Implied)<br>- Interpretable*, Traceable*, Explainable* (Claimed Qualitatively)<br>- Robustness via Generalizability (Usually Claimed Qualitatively) | - High-consequence, safety-critical<br>- Human-machine Teaming (ex., decision support in adversarial or safety-critical settings)<br>- Open-ended (i.e., dynamic and unpredictable) environments that will benefit from generalizability and learning in the field<br>- Multi-task solutions (can benefit from context-aware memory) |
| **More Efficient Use of Resources:**<br>- Data Efficiency (40-96% less data, 0-1% of labels)<br>- Computational Speedup (2-15,000x)<br>- Network Size (1/3rd parameters) | - Data-limited domains<br>- Tactical edge (including continuous learning at the edge)<br>- SWAP-constrained platforms |

## 2.3 ROADMAP TO ACCELERATE KNOWLEDGE-INTEGRATED INFORMED AI RESEARCH AND DEVELOPMENT FOR NATIONAL SECURITY

To quickly begin realizing the performance benefits that knowledge-integrated informed AI techniques promise to deliver to the national security domain, a first step is acknowledging "knowledge" as an important building block of AI technology. While other research directions are pursued to advance modern AI without knowledge integration, we should be considering which knowledge-bases should be integrated and how much importance the provided knowledge should be given over the standard approach.

Pursuing knowledge-integrated informed AI research for national security can come down to several deliberate steps that are data-oriented, knowledge-oriented, or that serve to advance the state of the art and its adoption into the national security domain. Figure 11 lays out some immediate steps we should be taking. In addition to realizing the performance advantages described in this section, all of these steps can be honed toward addressing existing challenges and pursuing promising yet untapped research opportunities (discussed later in Section 6)

# TABLE 4

**Rudimentary Examples of Knowledge-Integrated Informed AI for National Security**

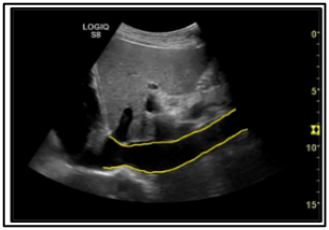

**Biotechnology**

Deep learning based ultrasound image interpretation guided by knowledge graphs → more generalizable

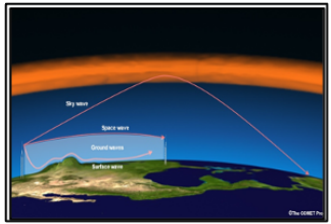

**Missile Defense**

Integrate knowledge of atmospheric effects on signal propagation into deep learning for signal classification/discrimination → greater accuracy

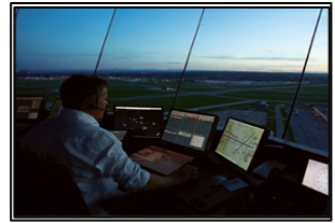

**Air Traffic Control**

Consider symmetries in reinforcement learning approach for air traffic collision avoidance → faster and more efficient learning

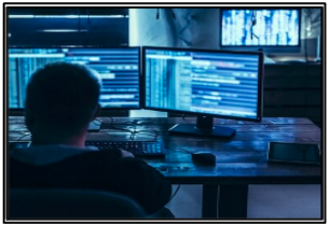

**Cybersecurity**

Consider symmetries in cyber-network-based neural network topologies to maintian connectivity and immunity to vulnerabilities → more efficient learning, more verifiable models

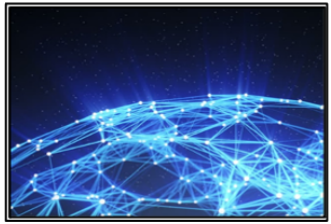

**Communications**

Integrate communication network access rules and limits into a deep learning based secure access and routing scheme → correctness-by-construction

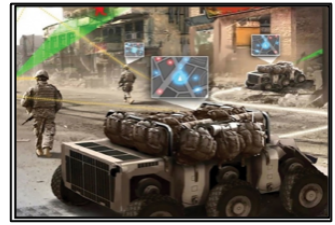

**Engineering**

Constrain autonomous control and planning algorithms with necessary rules of engagement → improved accuracy and validity of recommendations

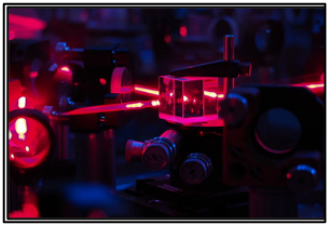

**Technology**

Use expert-specified examples and physical constraints to discover and recommend better experiment configurations → greater likelihood of valid recommendations

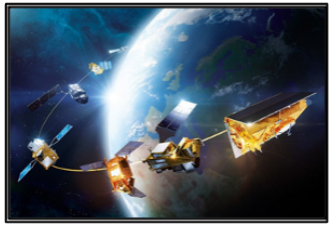

**Space Systems**

Use orbital mechanics equations to constrain optimal satellite control learning algorithm → greater accuracy and more generalizable

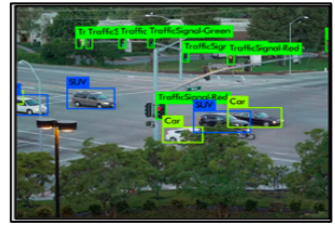

**Surveillance**

Exploit human implicit and explicit semantic knowledge (graphs) for reasoning over object categories → greater accuracy and more generalizable

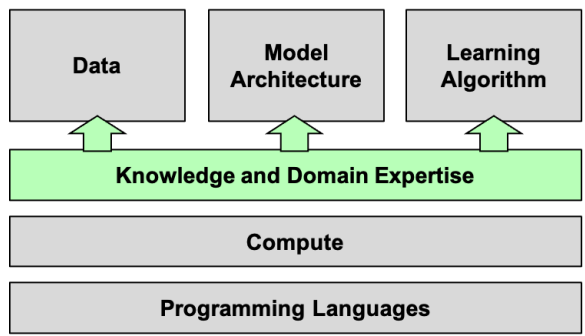

*Figure 10.* *Add Knowledge and Domain Expertise as an Important Building Block of AI Technology Knowledge/domain expertise should be recognized as important building block of AI technology for national security.*

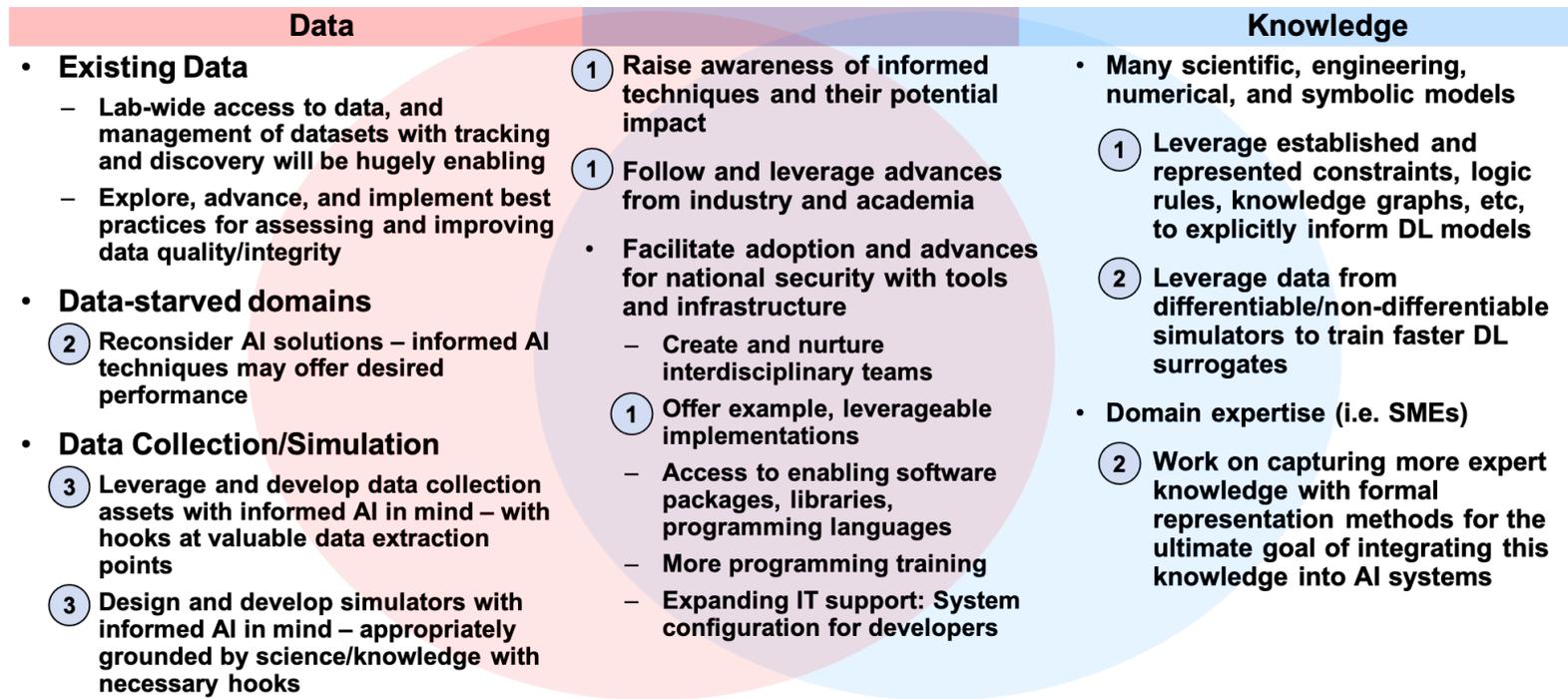

**Figure 11.** *Roadmap to Accelerate Knowledge-Integrated Informed AI Research and Development for National Security*
*Here we list several steps that will help accelerate knowledge-integrated informed AI research and development for national security that are either data-oriented, knowledge-oriented, or serving to advance the state of the art and its adoption into the national security domain. Circled one, two, and three indicate first, second, and third priority.*

This page intentionally left blank.

# 3. WHAT IS KNOWLEDGE-INTEGRATED INFORMED AI?

We mostly adopted Laura von Rueden et. al.'s definition of "informed machine learning" into our definition of "knowledge-integrated informed AI," which we presented earlier in Section 1 and repeat here. Note that we more exclusively focus on the integration of knowledge into more "modern" AI techniques, i.e., deep learning and reinforcement learning.

Laura von Rueden et. al.'s define "informed machine learning" in their 2021 survey paper [1] as the following: *Techniques that learn from a hybrid information source which consists of both data and prior knowledge.* In this definition,

- knowledge is "true" or "justified belief," validated either via science, empirical studies, or experts, where the more formally knowledge is represented, the more easily it can be integrated.

- the source of knowledge can be from the natural sciences, social sciences, expert knowledge, or world knowledge.

- prior knowledge is explicit, pre-existent, separate from the data, and integrated into the machine learning pipeline.

- informed deep learning and reinforcement learning are encompassed, and other classes of AI are not excluded (for example they include probabilistic graphical model based methods, like Bayesian networks).

Our adapted definition of knowledge-integrated informed AI is the following:

> **Knowledge-Integrated Informed AI**: AI techniques in which explicit, principled and/or practical knowledge, that is scientific, domain-specific, or data-centric, is integrated (not necessarily in a separable way) into a model's development pipeline. In these techniques:
>
> - Knowledge from scientific or domain expertise is integrated to provide partial explanations or constraints for the unknown relationship that the model is built to learn,
>
>   Or
>
> - Understanding of properties of training data are exploited to provide more efficient/effective mechanisms for learning.

## 3.1 ARCHITECTURES FOR INTEGRATING KNOWLEDGE

Before necking down into knowledge-informed AI pipelines, we should discuss some different architectures for knowledge integration. The integration of explicit knowledge into otherwise "uninformed" AI pipelines can be anywhere from loosely coupled to tightly integrated. Figure 12 illustrates the various system architecture we've encountered. For this study, we focused largely on tightly integrated architectures of informed deep learning and reinforcement learning—covered in Sections 3.3 and 3.4 with illustrating examples in Section 4 We also looked into embedded and jointly trained neural networks, which we cover in Section 4.5.

While also a worthwhile direction, since such approaches are more common and straightforward, we did not deeply investigate loosely coupled architectures. These consist of serial approaches where (1) a knowledge-based model could narrow the search space for a neural network, or (2) where a knowledge-based model refines or facilitates the predictions that come out of a neural network; and "complementary" configurations where there can be adaptive selection between knowledge-based or neural network models. In other words, we did not deeply explore neuro-symbolic AI that has been defined as "modular systems that seek to have the property of compositionality" [32].

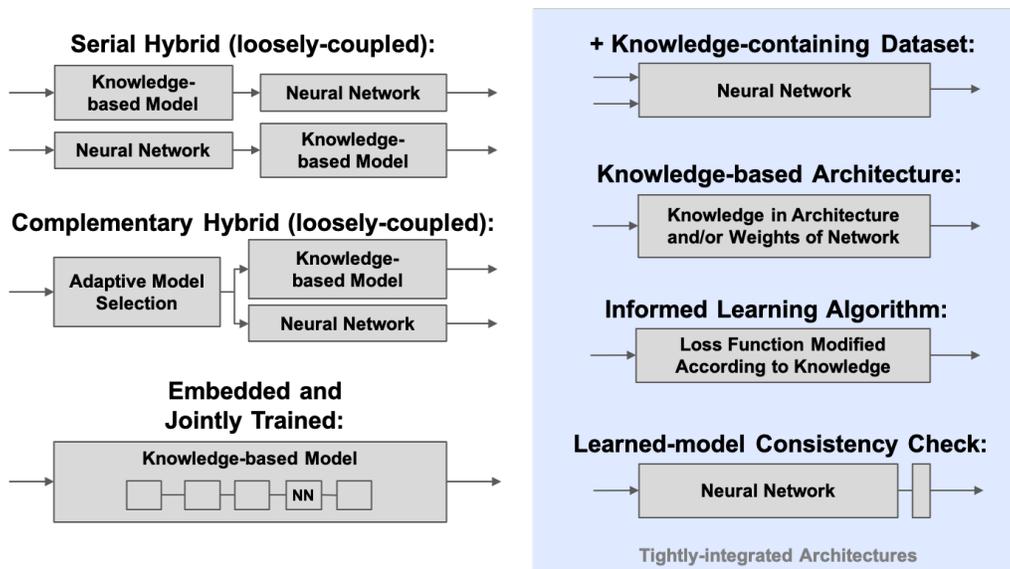

**Figure 12.** Loosely Coupled to Tightly Integrated Informed Deep Learning Architectures
Loosely coupled: Serial hybrid or complimentary hybrid configurations. Tightly integrated: additional knowledge-containing dataset (a.k.a., informed via training data), knowledge-based architecture and/or weights (a.k.a., informed hypothesis set), informed learning algorithm, learned-model consistency check (a.k.a., informed final hypothesis).

## 3.2 KNOWLEDGE REPRESENTATIONS

Illustrated in Figure 13, Laura von Rueden et. al. enumerate eight forms of knowledge representation [1] that can be integrated into informed AI pipelines. We mostly adopt their list making only one minor modification: renaming to "differential equations" to "dynamical system." Also, we indicate on their list which of the knowledge forms can have symbolic representations.

## 3.3 KNOWLEDGE-INTEGRATED DEEP LEARNING PIPELINE

In general, deep learning models are developed to learn unknown input/output relationships of a given problem. With features *in* and predictions *out*, the model is trained to fit the provided data. Laura von Rueden et. al. define a generic "informed machine learning" pipeline illustrated in Figure 14. This pipeline works for the new and emerging paradigm of knowledge-integrated informed AI that is based on deep learning techniques, however not perfectly, and we indicate why below.

| | |
|---|---|
| **Algebraic Equations** $E = m \cdot c^2$ $v \leq c$ | **Algebraic Equations: aka logic constraints**<br>• Represent knowledge as equality of inequality relations between mathematical expressions consisting of variables or constraints<br>• Equations can be used to describe general functions or to constrain variables to a feasibly set |
| **Logic Rules** $A \wedge B \Rightarrow C$ | **Logic Rules: aka logic constraints**  **Sometimes Symbolic**<br>• Provide a way of formalizing knowledge about facts and dependencies and allows for translating ordinary language statements (if/then) into formal logic rules<br>• Can be transformed from and to a knowledge graph |
| **Simulation Results** | **Simulation Results:**  **Sometimes Symbolic**<br>• Describe the numerical outcome of computer simulation – these include differential solvers and other simulators that incorporate complex forms of knowledge |
| **Differential Equations** $\frac{\partial u}{\partial t} = \alpha \frac{\partial^2 u}{\partial x^2}$ $F(x) = m \frac{d^2 x}{dt^2}$ | **Dynamical Systems* (differential eqns. and discrete models)**  **Sometimes Symbolic**<br>• Subset of algebraic equations that describe relations between functions and their spatial or temporal derivatives<br>• Often the basis of computer simulation |
| **Knowledge Graphs** | **Knowledge Graphs:**<br>• Nodes or vertices describe concepts and edges represent relations between them  **Always Symbolic** |
| **Probabilistic Relations** | **Probabilistic Relations:**<br>• Prior knowledge could be assumptions on the conditional independence or correlation between random variables or even a full description of the joint probability distribution  **Sometimes Symbolic** |
| **Invariances** | **Invariances:**<br>• Properties that don't change under mathematical transformation such as translations and rotations |
| **Human Feedback** | **Human Feedback:**  **Symbolic Reasoning**<br>• Technologies that transform knowledge via direct interfaces between humans and machines.<br>  - Human demonstration of problem solving<br>  - Human feature engineering or model steering via weight sliders<br>  - Human using visual analytics |

***Figure 13.*** *Forms of Knowledge Representation from Laura von Rueden et. al. [1] This is an excellent and comprehensive list of knowledge representations. *We modified this list slightly, renaming "differential equations" to "dynamical systems" so to also include discrete-time models—ex., Markov decision processes, state machines, etc.*

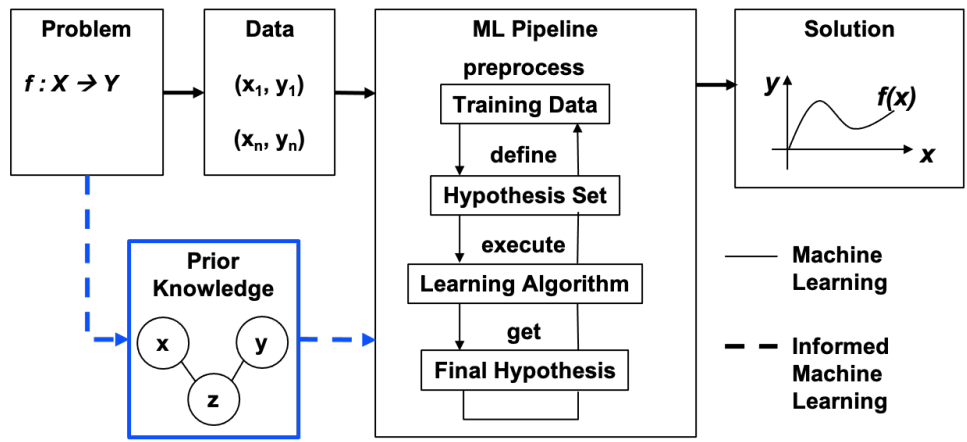

*Figure 14.* Laura von Rueden et. al. Informed Machine Learning Pipeline [1]

As drawn, the pipeline in Figure 14 does not clearly highlight what's different about this AI paradigm from standard approaches. Namely, this pipeline can be used also for approaches where knowledge is used to only implicitly inform or influence learning. For example, considering prior knowledge in the data pre-processing step is common practice that many refer to as feature engineering. However, implicit informing and standard feature engineering don't fit the textual definition (not ours nor Laura von Rueden's). Instead, new ways of *explicitly integrating* knowledge do fit—for example, leveraging additional knowledge-based datasets (though not the most promising knowledge integration path, more on this in Section 4).

In Section 4, we discuss the four established integration paths (training data, hypothesis set, learning algorithm, and final hypothesis) in more detail, along with examples that we found to be illuminating. We also share our overarching observations based on reported performance gains and the apparent state of the art.

As we discuss illuminating examples of knowledge-integrated informed AI variants, we populate a table of reported performance gains that is structured around the taxonomy of knowledge forms and integration paths. This table, presented in Section 5, allows us to extend the discussion from promising performance gains to which variants may be more suitable for different problems.

### 3.4 KNOWLEDGE-INTEGRATED REINFORCEMENT LEARNING PIPELINE

The generic machine learning pipeline described in the previous section is broadly relevant, beyond just deep learning, and even applies to reinforcement learning. However, while the forms of knowledge representations can largely remain the same, informed reinforcement learning has its own framework with unique integration paths.

Generally, reinforcement learning (RL) is a sub-field of machine learning where the goal is to train an agent to make optimal choices in a sequential decision making process. In other words, at any given moment, the agent should select the best action, taking into account the future effects of its decision. The action can come from a policy, which maps a state to an action; a value function, which maps a state to the expected future performance from that state; or a q-table, which maps a state-action pair to the expected future performance.

In standard (i.e., non-deep) RL, policies, value functions, and Q-tables are typically represented as lookup tables. These tables need an entry for every state or state-action pair. For many real-world systems, the size of these tables renders the learning process effectively intractable. Furthermore, all states (or state-action pairs) need to be explored in order for the agent to have a good idea of what the performance will be in those states. Entries in the policy, value table, or q-table do not generalize to other unvisited states.

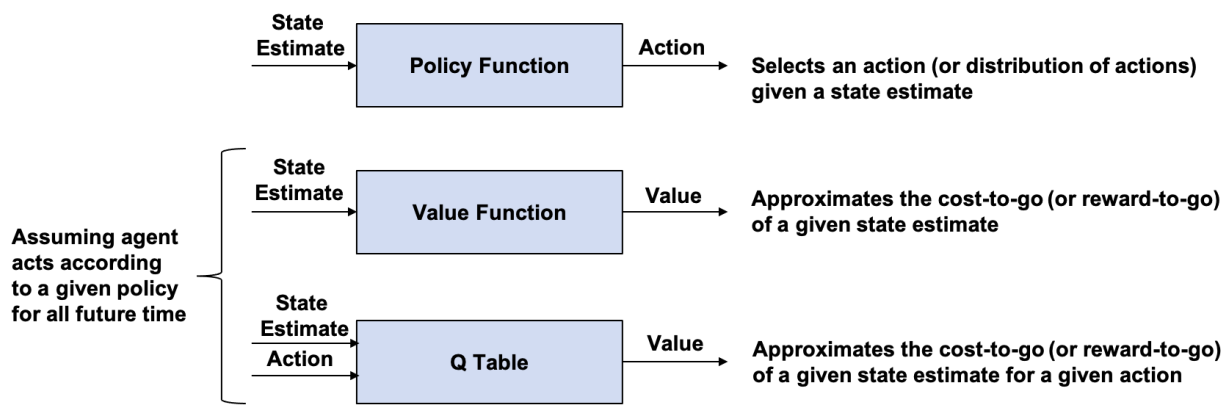

*Figure 15.* Components of Non-deep RL
*Components that are learned in non-deep RL setting include policy functions, value functions, and/or Q-tables. These functions are trained to return the best action from a given state, or the value of an action from a given state. These methods have issues of scalability and generalizability to states that were not seen during training.*

Because of these issues of intractability and generalizability, there is interest in deep RL where the policy, value table, or q-table is replaced with a neural network approximator. Now, instead of large tables, the learned function is a relatively small network. For example, in the case of a three-dimensional state space, with five possible values per state and five actions, the Q-table needs 625 entries. The corresponding neural network needs three input nodes (one per dimension) and five output nodes (one per action), and a small number of hidden layers. In concrete terms, this can lead to a significant reduction in the storage size of the policy. For example, in ACAS-X, the policy table is 1.22 GB, but a neural network approximation is 104 KB (Figure 17) [33].

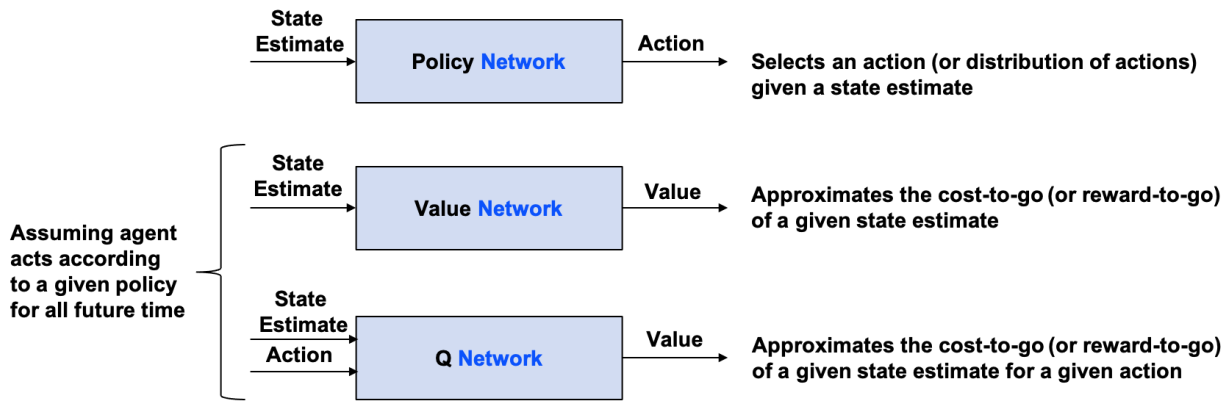

*Figure 16.* Components of Deep RL
Components that are learned in deep RL setting policy networks, value networks and/or Q networks. These networks approximate the tables and functions in Figure 15. They do not suffer from the same issues of scalability and generalizability encountered in non-deep RL.

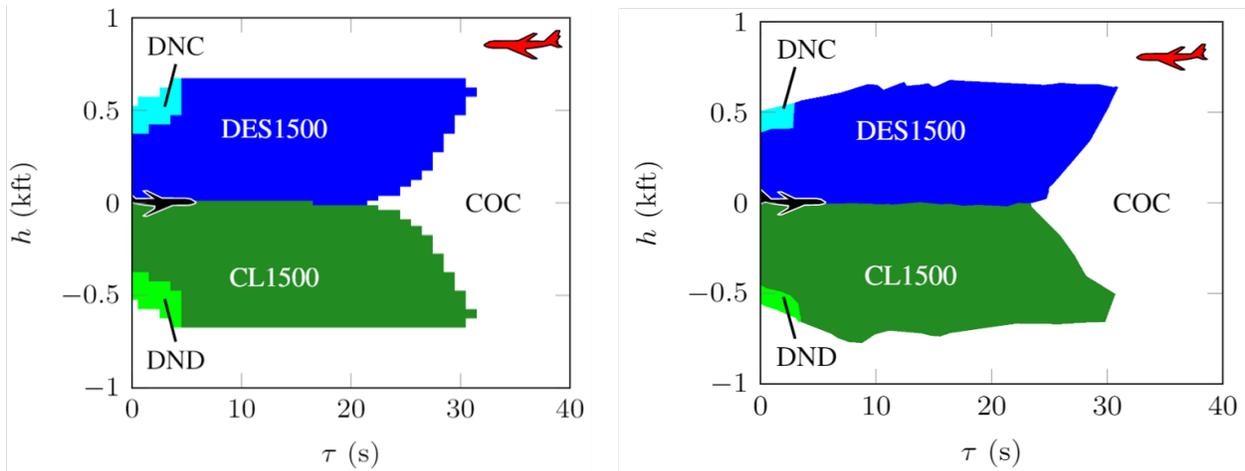

*Figure 17.* Q-Table vs. Neural Network
ACAS-X collision avoidance policy as a Q-Table (left) and a neural network (right). The neural network requires much less storage, but it may differ from the original policy. Images from [33]

A neural network representation can also generalize, assuming similar inputs should lead to similar outputs. Effectively, unvisited states get evaluated according to the "closest" state that has been visited. On the other hand, this generalization makes verification and other analysis harder for a neural network than a table.

The deep RL pipeline has essentially four main steps (Figure 18). First, the agent takes action according to its policy and observes the results transition, observations, and rewards. Second, the history is saved. The process of saving the history varies according to different algorithms (out of scope for this report). Third, a loss function is computed. Again, this process varies according to the specific algorithm under consideration, but the loss is often a property of the value network or Q-network (such as self-consistency according to the Bellman equation) or the difference between observed and expected performance. The loss is weighted by the accumulated reward. Fourth, the loss is backpropagated across the neural network that represents the policy, value network, or Q-network. These steps repeat until performance stabilizes, or some other termination criteria are reached.

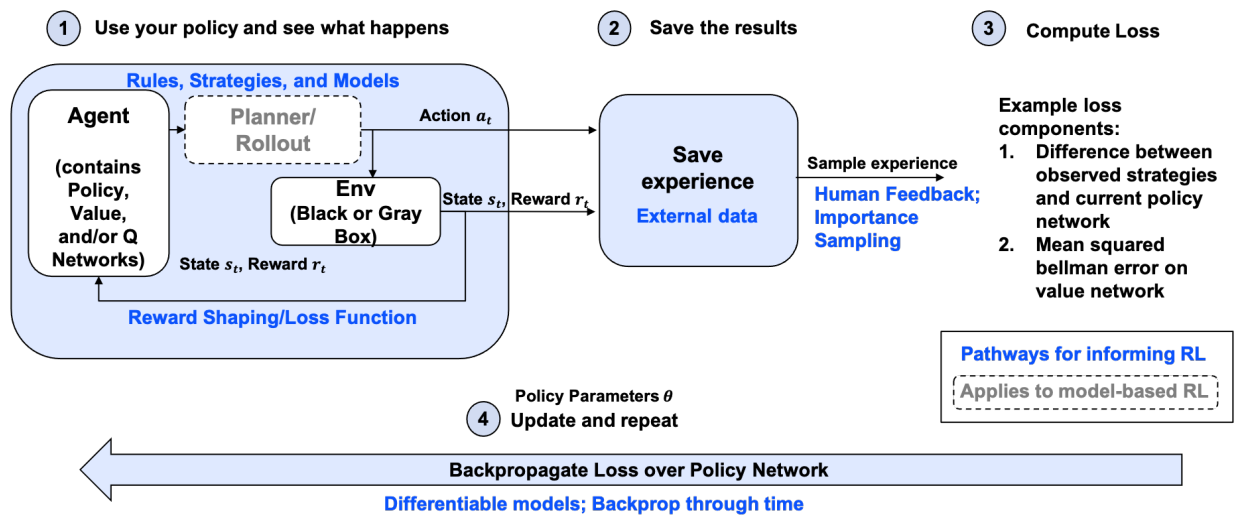

*Figure 18.* Deep RL Pipeline

The typical deep RL pipeline includes fours steps: 1) Executing a policy and observing the results, including transitions and rewards; 2) saving the results; 3) computing a loss function on the results, weighted according to the received reward; and 4) updating the policy via backpropagation. This process repeats until a termination condition is met. Different pathways for informing the RL pipeline are highlighted for each of the four steps in the pipeline. Blue text indicates pathways for informing the pipeline. The dashed box applies to model-based RL.

During the literature review performed for this report, we came across several main pathways for informing deep RL. These pathways are indicated in green in Figure 18. For step 1, reward shaping or modifying the loss function are common pathways for informed deep RL. Additionally, step 1 can also accommodate rules, strategies, and models that can be used by a planner to select actions based on a policy or value network. For step 2, external data, such as human gameplay or other demonstrations, can be added to the set of saved experience. In step 3, human feedback or importance sampling can determine what data is used, or how it is weighted, in the loss function. For example, human demonstrations could be sampled more frequently to force the RL process to mimic the human examples. In step 4, differentiable models can be used as part of the backpropagation process, sometimes termed "backpropagation through time." This type of backpropagation is sometimes accomplished using an LSTM. In subsequent sections of this report, we highlight examples of some of these pathways for informing deep RL.

We note that the pipeline we describe here, displayed in Figure 18, is geared specifically toward *deep* RL. However, as the reader might infer from Figures 15 & 16, the main differences between RL and deep RL lie mainly with how policies are stored and updated. Many of the other components in the learning pipeline are the same. Therefore, many of the pathways for informing the deep RL pipeline apply equally well to RL in the absence of deep learning.

# 4. ILLUMINATING EXAMPLES

## 4.1 KNOWLEDGE-INTEGRATED VIA ADDITIONAL KNOWLEDGE-CONTAINING DATASET(S)

Explicitly integrating knowledge into training data comes down to augmenting an original data set with an additional, (separate) dataset that captures important patterns with some certainty. Figure 19 aptly captures this concept.

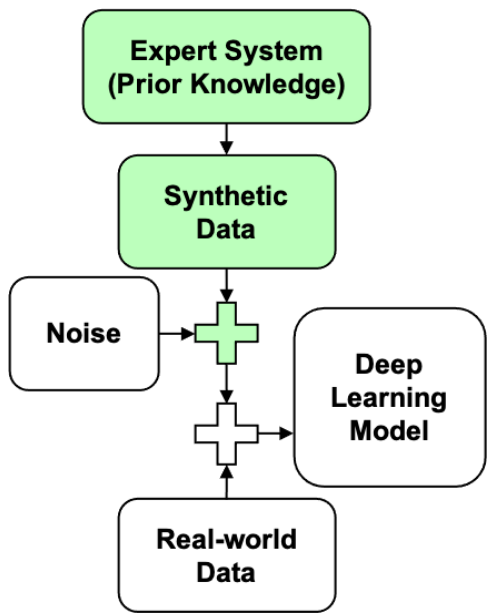

**Figure 19.** *Architecture with Additional Training Data from [2] (Aug 2018)*

*This broadly applicable approach of leveraging prior knowledge through synthetic data generation is described nicely in "Learning from the experts: From expert systems to machine-learned diagnosis models" [2] (Aug 2018).*

The additional dataset(s) often, but not necessarily, come from simulations (that can leverage understanding of the domain in the form of algebraic equations, knowledge graphs, or invariances), or via expert demonstration of the task at hand (Figure 20). For example:

- When knowledge is represented in the form of algebraic equations or knowledge graphs, it can be leveraged to train data generators [34] such that the generated data is then used to supply samples into a widened neural network input layer.

- Additional data may also come from having access to additional information, such as access to a simulation's background processes [35]. These kinds of data augmentation have been shown to improve accuracy and generalizability.

- Likewise, supplementing with data generated by human demonstrations of performing the task at hand (that the neural network is being trained to perform) also has been shown to result in improved accuracy [36].

- Claims of greater accuracy and generalizability along with data and computational efficiencies come from integrating knowledge of invariances (i.e., symmetries across mathematical operations, such as translation, reflection, rotation, and scaling of an image [37], which may be used to create and use new, virtual examples that are transformations of samples in original datasets [37] [38].

Many of the examples listed above may blur as, simply, "feature engineering," but the knowledge-based data source(s) being explicit or separate from other data is a distinguishing factor. Nonetheless, the performance gains reported from integrating knowledge into deep learning pipelines via additional training data touch several important metrics: accuracy, generalizability, data, and computational efficiencies. Performance in terms of accuracy is sometimes claimed only qualitatively, but when quantified we saw gains from 1-7%, 0.6-1.7%, and 2.4%, for example. These gains, when compared to other knowledge integration paths, may be considered marginal to moderate, however may be be sufficient for certain applications. Improvements in generalizability and data or computational efficiencies may also be worthwhile gains that make all the difference.

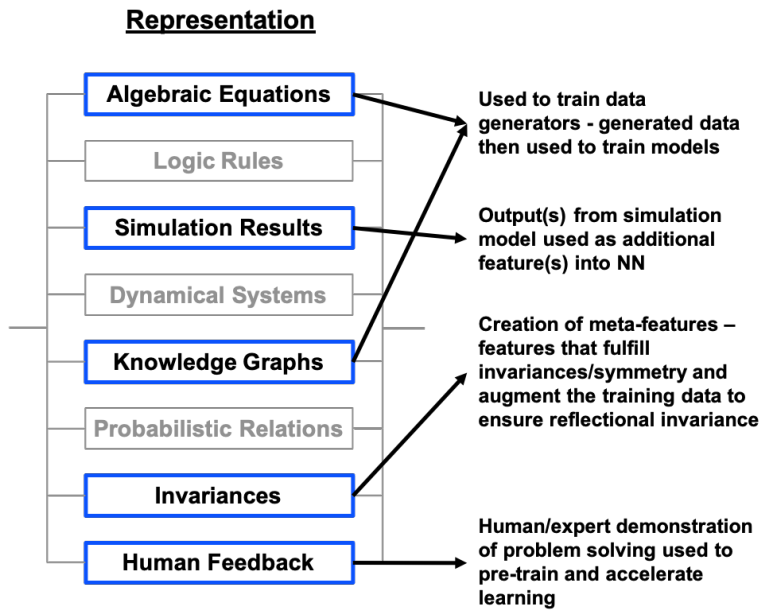

**Figure 20.** *Forms of Knowledge Shown to Integrate via Training Data [1]*

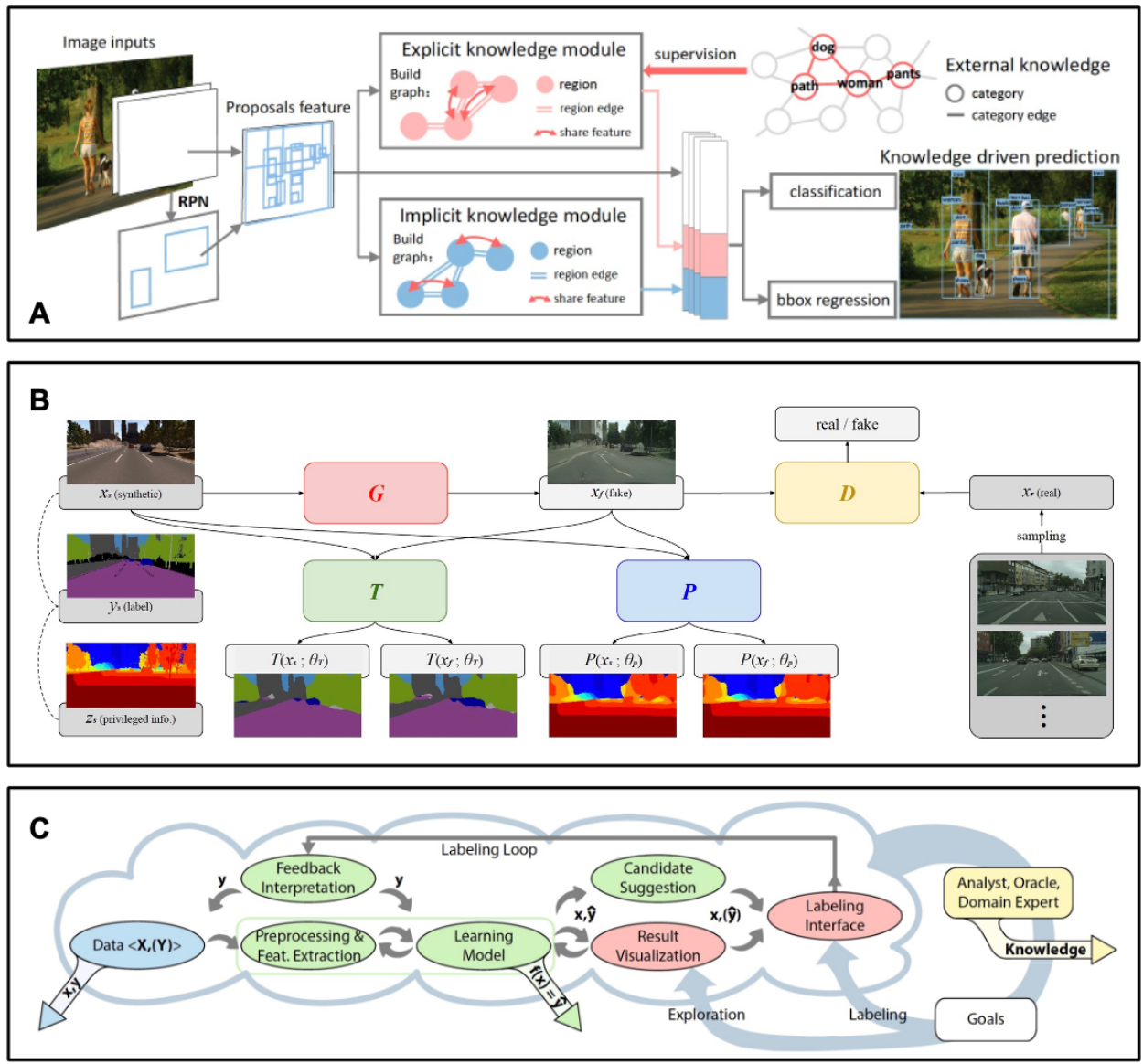

*Figure 21.* *Examples from Literature that Illustrate Knowledge Integration via Training Data PANEL A, from [34], illustrates data augmentation from both explicitly and implicitly defined knowledge graphs. PANEL B, from [35], shows four networks (G-generator, D-discriminator, T-perception task, and P-prediction of simulator's privileged information) that are trained jointly to improve the perception task's performance. PANEL C, from [36], illustrates the active learning approach where a human analyzes model results in order to suggest new instances for data labeling.*

### 4.1.1 Integrating Knowledge via Training Data for Informed Reinforcement Learning

For reinforcement learning, it has been shown that simulation results can be leveraged to augment the main learning loop. Inherently, reinforcement learning involves executing actions in the real world or in simulation. However, additional information can be provided via simulation for use during online adaptation. For example, in "Robots that can adapt like animals" [39], the authors build a low-dimensional behavioral map of robot performance in simulation during the initial training phase (Figure 22, top row). The low-dimensional map reduces the high-dimensional state space of the robot to a behavioral map that measures expected performance along different behavioral axes (e.g., speed, heading, etc.). At run time, the robot can compare its actual performance to its expected performance via the behavioral map (Figure 22, bottom row). If the performance degrades when the robot is deployed, due to damage or other issues, the robot will identify the discrepancy between its expected performance and the behavioral map. It will then update the map and attempt a new strategy from the updated map.

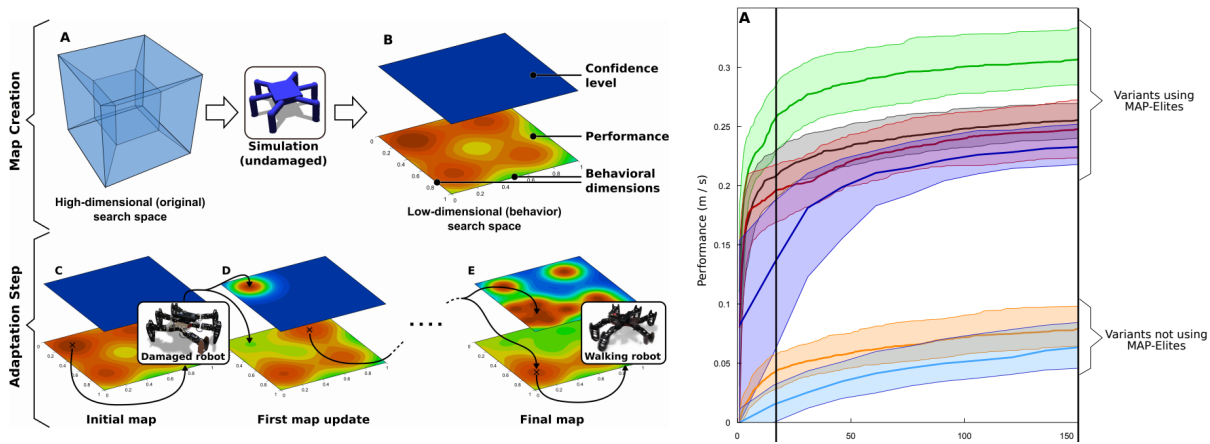

*Figure 22.* RL Example Informed via Additional Simulated Data
(LEFT) The two steps in using a performance map to improve online learning to recover from damage. Top row: creating an initial low-dimensional performance map from simulation data. Bottom row: updating the map online and using it to sample new behaviors. (RIGHT) Comparison of algorithm performance on a damaged hexapod. Bottom blue and orange curves represent uninformed RL approaches. The top set of curves represent variants of the algorithm presented in the paper. [39]

This efficient sampling and updating of simulated data with real-world experience can lead to significant performance improvement in reinforcement learning applications. The results of this process for a hexapod robot led to approximately 3-5x performance improvement over uninformed RL as measured by the walking speed of the robot after sustaining damage (Figure 8). RL is in general a resource intensive process, requiring significant training experience to get good performance, and it can struggle to scale with high-dimensional spaces. Informing RL with this type of simulation data aids the process by providing an expectation of performance a priori (thereby reducing the amount of experience that needs to be generated online), and reducing the dimensionality of the search space (thereby reducing the difficulty of efficiently sampling the search space).

## 4.2 KNOWLEDGE-BASED ARCHITECTURE OR WEIGHTS (A.K.A., HYPOTHESIS SET)

Our literature review included many creative examples of integrating knowledge into the hypothesis set, i.e., the initial architecture and/or weights of a neural network. Collectively, the examples we reviewed indicate that this integration path can often deliver significant improvements in performance. We found that the degree of knowledge integration into neural network architectures can range from modifying the input layer (via invariances, knowledge graphs), to constraining intermediate layers (via algebraic equations, logic rules), to influencing the overall framework (using simulation results, probabilistic relations, or dynamical system constructs). Human feedback by model steering (via weight sliders, choice of model, etc.) is also considered a means of integrating knowledge into the neural network hypothesis which can be useful in applications where valuable knowledge may not be as readily and formally represented. For example:

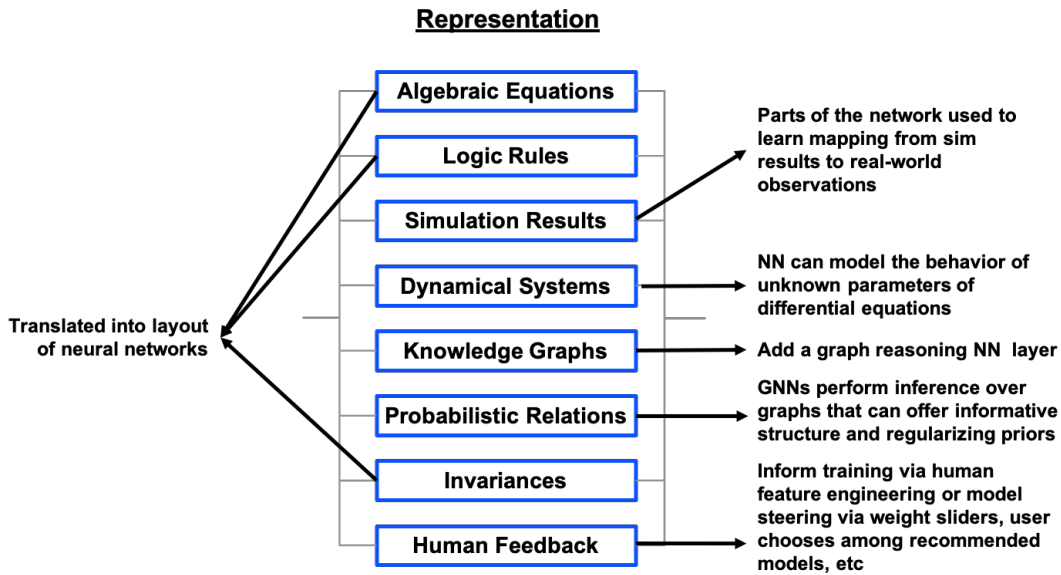

*Figure 23.* Forms of Knowledge Shown to Integrate via Architecture and/or Weights [1]

- To account for known invariances, additional input layers can be embedded into an architecture [40] [41] (panels A and E in Figure 24). Note that embedding invariances into a neural network architecture has been reported to yield both a larger network (1.5x [42]) as well as a smaller network (1/3x [41]).

- Alternatively, a neural network can be architected to allow "dynamic" input vectors with zero padding that is based on importance and determined, for example, via reasoning through a knowledge graph [43] (panel B in Figure 24).

- Moving beyond the input layer, physics-based constraints can be enforced by mathematically derived (via algebraic equation) intermediate variables in a neural network [44] (panel C in Figure 24).

- Graph neural networks (GNNs) have a unique architecture that allows them to directly process graphical data (including probabilistic graphs) where knowledge may be present in the graph structure and/or priors. GNNs have been shown to offer greater efficacy in handling loopy graphs. They also deliver greater accuracies, and generalizability (than graphs that are larger or with less "informed" structure) [45] [46] (panel F in Figure 24).

- More holistically, a neural network can be architected to learn unknown parameters in differential equations, or other mathematical frameworks, such that the architecture is determined by the necessary mathematical constructs [47] (panel D in Figure 24).

Deep Lagrangian Networks (DeLaN from [47]) is a particularly elegant example where principled knowledge was infused into the deep learning system. Both the model's architecture and the gradients used to train it are derived from a rich understanding of mechanics (Euler-Lagrange equations of motion for mechanical systems). In this example, the aim is to solve a second order ordinary differential equation (ODE) by being trained to infer, from generalized positions, velocities, and accelerations of a mechanical system's components. The DeLaN architecture explicitly incorporates the structure of the Euler-Lagrange equation and is designed specifically to relate a mechanical system's state with a control input (e.g., motor torques) to predict the system's evolution–outputs of the feed-forward neural network. The rest of the architecture incorporates these quantities with the observed system state to produce the generalized forces, via exact Euler-Lagrange calculations. The full architecture is trained in a traditional way, where the loss function effectively optimizes the DeLan's parameters to minimize the violation of Lagrangian mechanics.

In terms of effectiveness, the DeLan model effectively learns to relate system dynamics to produce joint torques for desired motion trajectories—directly attributable to the physics-informed design of its architecture. In terms of performance, not only does the learned model come with a high degree of interpretability and ability to handle arbitrary velocities and accelerations that fall outside of the training data, it reached this state using less data. These gains in performance are not quantified, but are nonetheless significant.

Generally, integrating knowledge via the hypothesis set into deep learning pipelines has been shown to deliver significant performance gains in terms of accuracy, generalizability, efficiencies in data, compute, and model size, and interpretability. Most of these improvements are reported qualitatively, but we found improvements in accuracy ranging from from 0.6-16%, reductions in test error on the order of -6 to -70%, requiring 40-57% less data, providing similar performance with a fraction (1/3rd) of network parameters, and 440x speedup in training time.

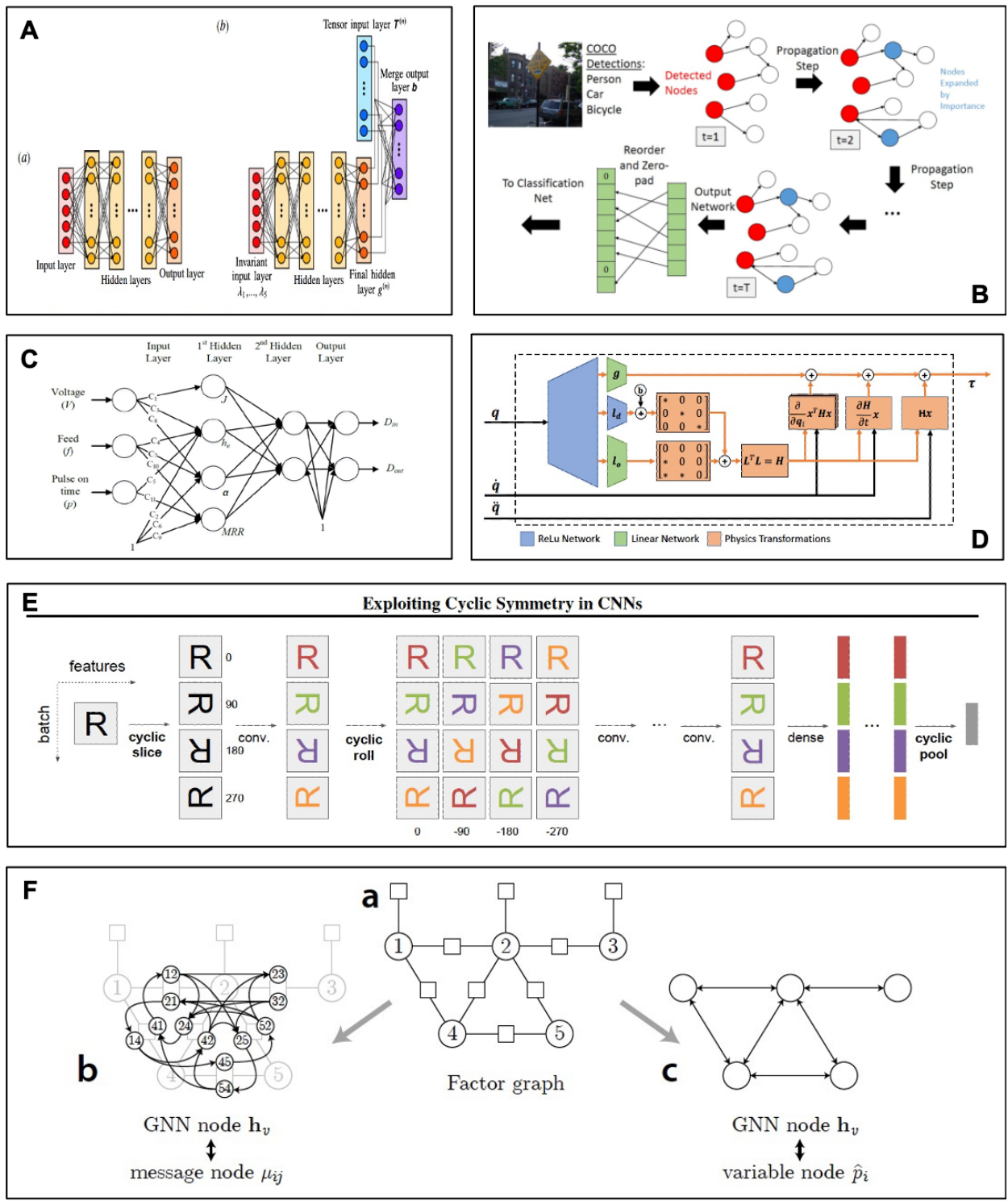

*Figure 24.* Examples from Literature that Illustrate Knowledge Integration via Hypothesis Set
PANEL A, from [40], shows how invariances deduced from Reynolds Navier-Stokes equations can be embedded into a neural network architecture. PANEL B, from [43], illustrates how weights/zero-padding in a network can be determined from propagation through a knowledge graph. PANEL C, from [44], illustrates physics-constrained intermediate variables. PANEL D, from [47], shows how physics-based, differentiable transformations can be built into a neural network. PANEL E, from [41] demonstrates how simple cyclic rotations are exploited in convolutional neural networks. PANEL F, from [46], illustrates a probabilistic graphical model is translated so that it can be processed by a graph neural network.

### 4.2.1 Integrating Knowledge via Hypothesis Set for Informed Reinforcement Learning

The reward function is a critical architectural element that directly effects the performance of reinforcement learning algorithms. It is the primary mechanisms for conveying the quality of an action or policy to the learning algorithm. For many tasks, designing a good reward function is hard, especially for tasks that are complex, have many steps, or have delayed rewards. With typical, Markovian, state-dependent rewards, accomplishing these types of tasks can require very long training time. Using logical rules and automata that encode knowledge about the structure of the task to inform the reward function has strong potential to address these difficulties.

The authors of the paper DeepSynth [48] present a method for informing the reward function using logical rules based on simulation results. At run time, a semantic segmentation module observes the scene and the reward (Figure 25, left). By identifying the semantic data corresponding to the reward, the algorithm can build an automaton capturing the reward (Figure 25, right). The only rewards specified by a human are very sparse—a reward of 100 for finding a key, and a reward of 300 for opening the door. This sequential task with sparse rewards is very difficult for most RL algorithms, but providing a reward automaton to the algorithm allows the agent to reason about the tasks that will lead to higher rewards in the future, resulting in convergence two orders of magnitude faster than the next best algorithm (Figure 26).

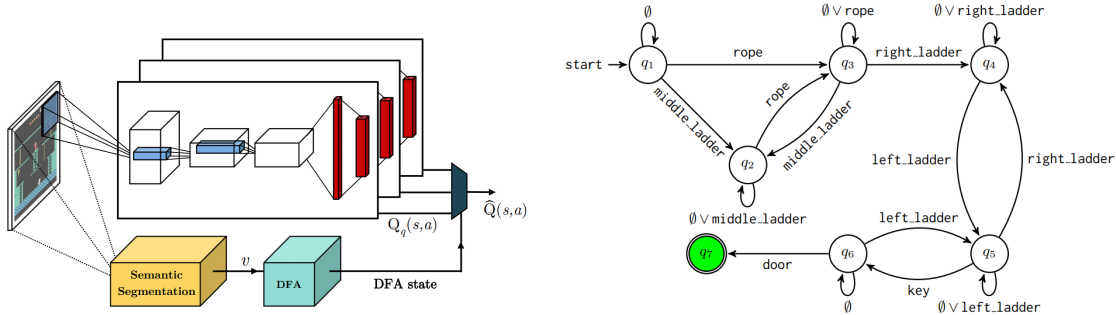

**Figure 25.** *RL Example Informed via Architectural Elements*
*Left: A semantic segmentation module captures information about the task during training and converts it to a reward automaton. Right: The reward automaton synthesized from Montezumas Revenge captures the high-level steps involved in playing the game. [48]*

This logical pathway for informing the reward has several other benefits. First, in addition to an automaton that is automatically constructed, it can also accept an automaton or partial automaton provided by the user. Rather than, or in addition to, specifying a sparse reward, a user can provide high-level tasking as an automaton. This automaton allows the agent to better understand the operator's goals. Further, the policy representation conforms to the automaton. Therefore, the operator gets the added benefit of interpretable policies. Thus, informed rewards

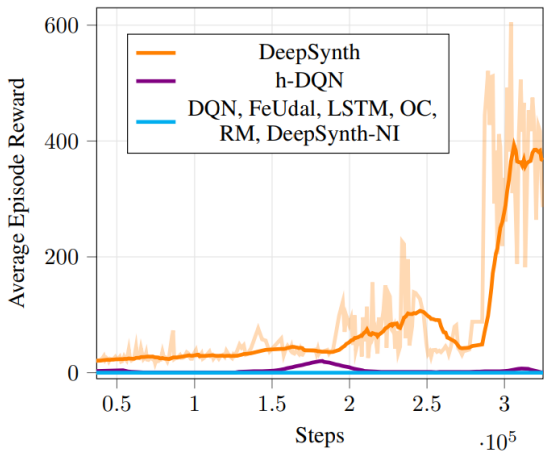

*Figure 26.* *Knowledge-Integrated RL via Architectural Elements - Performance*
*The methodology proposed in DeepSynth suggests that reward automatons that capture high-level aspects of the task outperform other RL methods and converge two orders of magnitude faster. [48]*

offer benefits to the algorithm and to the user. A user can specify non-trivial tasks that are difficult to express, resulting in much faster convergence, and the algorithm provides output that also informs the user about its goals and its policy.

### 4.3 KNOWLEDGE-INTEGRATED LEARNING ALGORITHM

A typical neural network supervised learning approach involves a loss function that aims to minimize the prediction error produced by the function $f_\theta(x)$ by finding optimal weights, $\theta$, for each sample, $x_1...x_N$, using real labels, $y$, as in Equation 1, where $R$ may be any standard regularizing term. So, integrating knowledge into the learning algorithm generally amounts to adding an additional term, or terms, to this loss function as in Equation 2. Note that additional parameters, as in Equation 3, may be included to control how steep the penalty should be for violating any of the terms in the loss function.

$$\arg\min_f \left( \sum_{i=1}^{N} L(f_\theta(x_i), y_i) + R(f_\theta) \right) \qquad (1)$$

$$\arg\min_f \left( \sum_{i=1}^{N} L(f_\theta(x_i), y_i) + R(f_\theta) + L_k(f_\theta(x_i), y_i) \right) \qquad (2)$$

$$\arg\min_f \left( \sum_{i=1}^{N} \lambda_l L(f_\theta(x_i), y_i) + \lambda_r R(f_\theta) + \lambda_k L_k(f_\theta(x_i), y_i) \right) \qquad (3)$$

According to the survey paper and our own literature review, it has been shown that several of the forms of knowledge representation can be integrated into the learning algorithm, or conversely, that the learning algorithm can be used to engage human expertise. For example:

- We found that additional knowledge-based loss terms, defined by algebraic or differential equations, have a straightforward path into the learning function and offer significant improvements to performance via improved accuracy or reduced test error, computational speedup and data efficiency [49] [50] [51] (examples B, C, and D in Figure 28). Note that if knowledge is represented as logic rules, appropriate continuous and differentiable constraints need to be selected for the additional loss term(s) [52].

- If knowledge is represented with a knowledge graph, the additional terms can effectively enforce strongly connected variables to behave similarly in a model [53]. Example A in Figure 28 claimed greater interpretability with their approach which led to the discovery of new clinically-relevant patterns.

- If knowledge is captured in a simulation or in auxiliary models, the learning algorithm can be modified to ensure that only predictions that obey constraints that are implied or embedded in those models effect the neural network learning. [54] illustrated in Figure 29 is an example that suggests that using semantic knowledge through regularized objectives can lead to slightly greater improvement in performance (overall accuracy and better generalizability) than additional inputs. We did not come across any reporting of the effect of integrating auxiliary tasks on compute time. Another example, [55], employed a similar approach by jointly running a physics-based simulation during training. This example claimed a reduction in the amount of data required for training along with improvement in accuracy and generalizability (but led to significantly more expensive compute time in simulation).

- Alternatively, a human can exploit the loss function using visual interactive analysis techniques. One approach is described in [36], where the training process using the standard loss function supplies samples on which the model performs poorly, and similar data samples are gathered and labeled for additional training–effectively focusing the model to train on its weaknesses. Another approach is described in [56] where a penalty is imposed when samples are classified correctly but for wrong reasons–thereby encouraging the model to focus on relevant features.

Overall, integrating knowledge into deep learning via the learning algorithm has been demonstrated to provide significant improvements across several important performance metrics: greater accuracies (+1.5 to +20%, or -4 to -43% test error), better interpretability and generalizability, data efficiency from requiring less data (as little as 4%) or less labeled data (1% to no labels at all), and computational speedups (4x).

### 4.3.1 Integrating Knowledge via Reinforcement Learning Algorithm

In some problems that are difficult for machine learning, humans are able to make decisions that are "good enough," without much difficulty. Often, this is because humans understand the rules of the game they are playing, or the way that other players will react to them. Imbuing an RL agent with this same understanding can significantly improve the performance of the RL algorithm. When actual human experience is included, the system has the potential to quickly match or even exceed human performance.

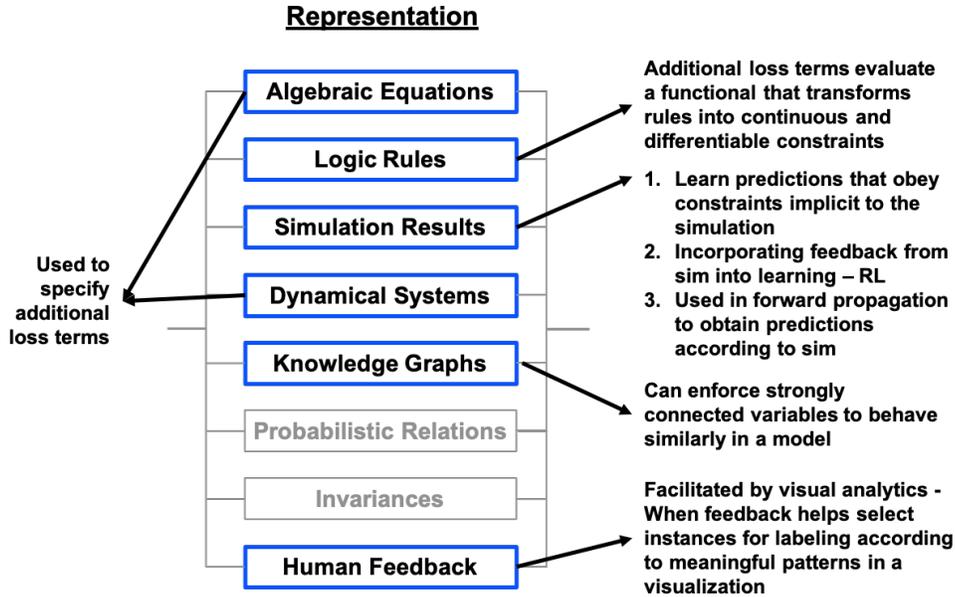

*Figure 27. Forms of Knowledge Shown to Integrate via Learning Algorithm [1]*

One example of integrating knowledge into the RL algorithm in this way is Alpha Go [4]. This approach includes models of the evolution of the game in the RL algorithm, along with human gameplay data, to develop a superhuman player for the game of Go. First, human expert data is used to initialize a policy network, which can then be used as the initial policy network by the RL agent (Figure 30). This network is then refined through self-play to improve over the human baseline. A value network is trained by comparing expected performance against actual performance using Monte Carlo tree search (Figure 31). This comparison is enabled by both a dynamical system representation of the game's dynamics (for determining the possible evolutions of the game), as well as the human data (for initializing the probability of moves by the other player). As a result, the Alpha Go algorithm was the first AI system to beat a professional-level human player in Go, and ranked 1.6 times higher by ELO (a common ranking system for comparing two players in zero-sum games) than the next best AI for Go.

This example demonstrates the value of incorporating knowledge into the algorithm, as well as human expertise directly in the training data. The state space of Go is extremely large, so a complete enumeration of the state space is intractable. By leveraging the models of the evolution of the game, and using probabilities from human gameplay and self-play, the system can develop useful internal models of how the system will evolve. It can measure the difference between the expectations output by its models and how the game unfolds to improve performance. Likewise, by

**A** **Source: Probabilistic/Knowledge Graph – Prior Based Regularization**
**Deep computational phenotyping**
**Interpretable and led to new discoveries**

Deep neural network with regularization on categorical structure:

$$\mathcal{L} = -\sum_{i=1}^{N} \log p(\mathbf{y}_i|\mathbf{x}_i, \boldsymbol{\Theta}) + \lambda R(\boldsymbol{\Theta}) + \frac{\rho}{2} \text{tr}(\boldsymbol{\beta}^\top \mathbf{L} \boldsymbol{\beta})$$

Where a graph Laplacian regularizer enforces the parameters β_k and β_k' to be similar:

$$\text{tr}(\boldsymbol{\beta}^\top \mathbf{L} \boldsymbol{\beta}) = \frac{1}{2} \sum_{1 \leq k, k' \leq K} A_{k,k'} \|\boldsymbol{\beta}_k - \boldsymbol{\beta}_{k'}\|_2^2,$$

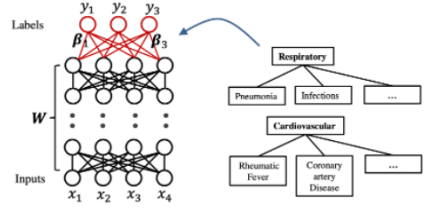

**B** **Source: Algebraic Equations**
**Incorporating prior domain knowledge into deep neural networks Incorporate loss terms for knowledge available as monotonicity constraints**
**4%-42.7% Less error**

$$\underset{f}{\text{argmin }} \text{Loss}(Y, \hat{Y}) + \lambda_D \text{Loss}_D(\hat{Y}) + \lambda R(f)$$

$$\text{Loss}_D(\hat{Y}_1, \hat{Y}_2) = \sum_{i=1}^{m} \mathbb{I}\left((x_1^i < x_2^i) \wedge (\hat{y}_1^i > \hat{y}_2^i)\right) \cdot ReLU(\hat{y}_1^i - \hat{y}_2^i)$$

**C** **Source: Algebraic or Differential Equations**
**Deep learning for physical processes: Incorporating prior scientific knowledge**
**Improved accuracy (-30% error) and faster run time**

The proposed NN model has been designed according to the intuition gained from general background knowledge of a physical phenomenon, here advection-diffusion equations. Additional prior knowledge – expressed as partial differential equations, or through constraints – can be easily incorporated in our model, by adding penalty terms in the loss function. As the displacement $w$ is explicitly part of our model, one strength of our model is its capacity to apply some regularization term directly on the motion field. In our experiments, we tested the influence of different terms: divergence $\nabla \cdot w_t(x)^2$, magnitude $(\|w_t(x)\|)^2$ and smoothness $\|\nabla w_t(x)\|^2$.

$$L_t = \sum_{x \in \Omega} \rho(\hat{I}_{t+1}(x) - I_{t+1}(x)) + \lambda_{\text{div}} (\nabla \cdot \hat{w}_t(x))^2 + \lambda_{\text{magn}} \|\hat{w}_t(x)\|^2 + \lambda_{\text{grad}} \|\nabla \hat{w}_t(x)\|^2$$

Weighted penalties: divergence, magnitude, smoothness

**D** **Source: Differential Equations**
**Physics-Informed Deep Generative Models**
**Data efficiency and computational speedup**

Final training objective encourages the generated samples to satisfy a given PDE read as:

$$\min_{\theta, \phi} \mathcal{L}_\mathcal{G}(\theta, \phi) + \beta \mathcal{L}_{\text{PDE}}(\theta)$$

*Figure 28. Examples from Literature that Illustrate Knowledge Integration via Learning Algorithm* Excerpts from several papers illustrate different modifications of the neural network loss function. In PANEL A, from [53], the learning algorithm encourages neural network predictions to be similar to the matrix representation of the probabilistic/knowledge graph. In PANEL B, from [49], an additional loss term forces monotonicity among specific input features. In PANEL C, from [50], the prediction loss is minimized while considering additional physics-based terms. In PANEL D, from [51], not satisfying a partial differential equation (PDE) encourages a generator (G) to produce samples that satisfy the PDE.

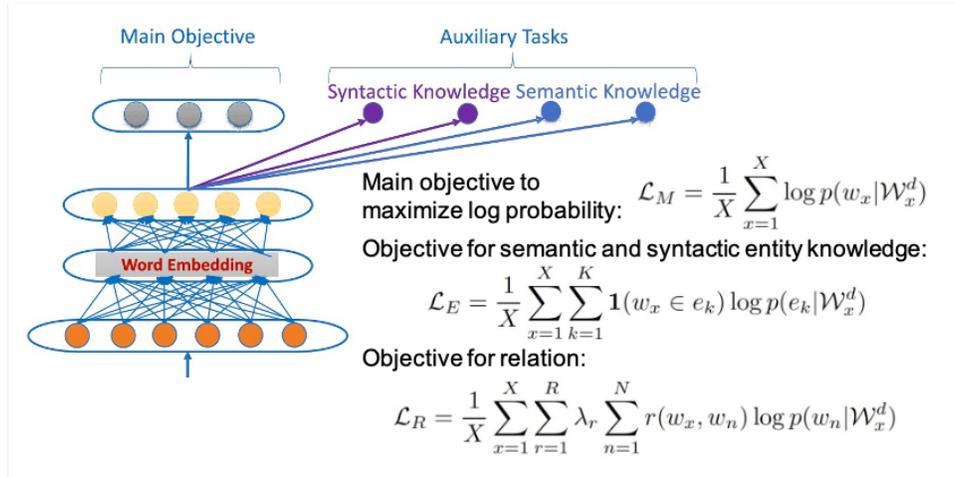

*Figure 29.* An Example from Literature that Illustrate Integration of Auxiliary Tasks via Learning Algorithm
The main objective $L_M$ in this example from [54] is to predict the center word given surrounding context. Additional loss terms, $L_E$ and $L_R$, incorporate semantic and syntactic knowledge via auxiliary tasks that leverage an entity vector and a relation matrix.

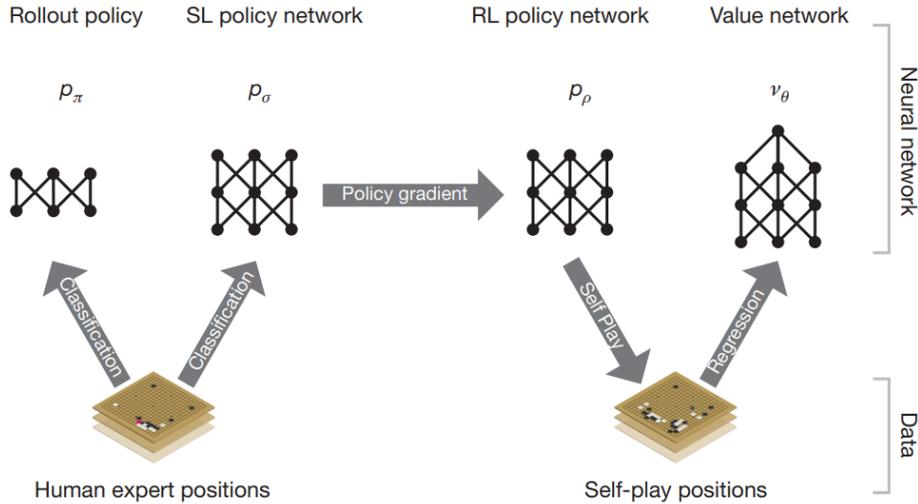

*Figure 30.* RL Example with Knowledge Integrated via Learning Process
LEFT: Human expert data is used to train a network to sample actions (rollout policy) and a network to play the game (SL policy network). RIGHT: The network trained on human gameplay is used to initialize an RL policy network that also learns a value network for the game via self-play. [4]

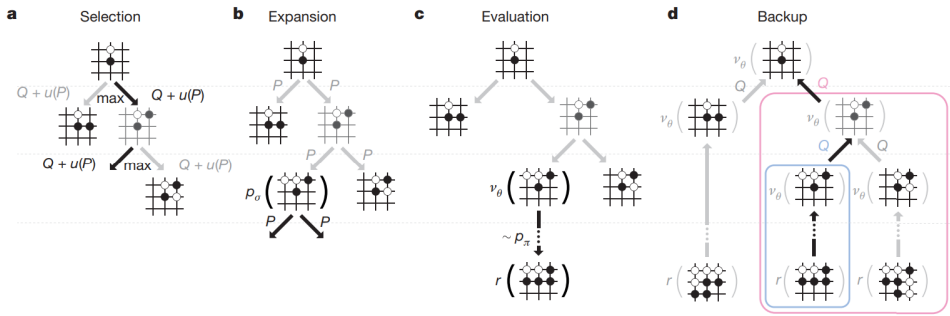

*Figure 31.* RL Example with Knowledge Integrated via Learning Process - Monte Carlo Tree Search
Monte Carlo tree search is used to evaluate moves based on game play. First, an action is selected with the maximum Q value (a). Then actions are sampled according to the policy network (b). The learned value network and rollout policy network (trained on human gameplay) estimate the outcomes of the game (c). Finally, Q values are updated along the tree. [4]

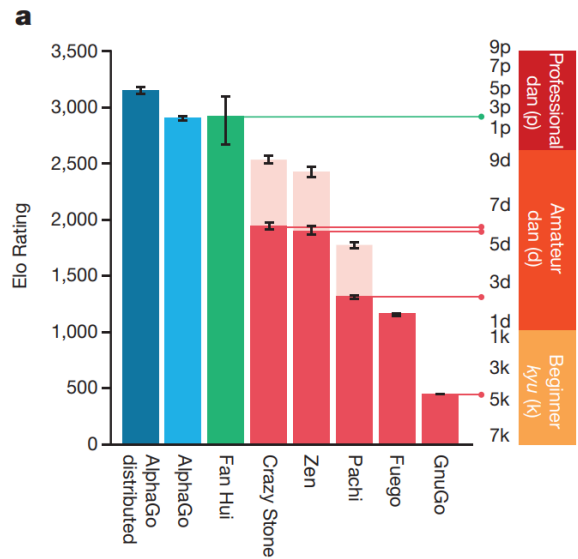

*Figure 32.* RL Example with Knowledge Integrated via Learning Process
Alpha Go (blue) was compared against several other Go programs (red) and a human expert (green). Distributed Alpha Go outperformed the human player, and both Alpha Go variants significantly outperformed the other Go programs, even with handicaps (pale red bars). [4]

including human expert data, the search space is further limited, and a more intelligent valuation of moves is possible. It also provides a warm start to the learning process. Instead of learning from scratch, the algorithm can learn from a good policy based on what expert humans perform. Informing an algorithm in this manner should lead to a significant reduction in training time. Without this information, the system would have to essentially luck into winning positions, and then reinforce those strategies. It should be noted that recent research has focused on achieving similar or better results without using problem-specific algorithms or human data. For more information on those research directions, see Sec. 1.2.2.

## 4.4 LEARNED-MODEL CONSISTENCY CHECK (A.K.A., FINAL HYPOTHESIS)

Generally, the aim of training a deep learning model is to arrive at a final hypothesis, after pursuing intermediate hypotheses, for the model and its weights. As such, there is opportunity to integrated knowledge into the learning pipeline during training. Knowledge is often leveraged as standard practice-when the performance of deep learning models is compared to selected baselines which are knowledge-based models (such as algebraic equations or simulation results). Here we shed light on a how knowledge may be leveraged by refining a model, as a final step or an integrated intermediate step in the learning process. As knowledge-integrated deep learning gains in popularity, we expect approaches that integrate knowledge at the final stages will too as they have been shown to yield greater accuracy, generalizability, and computational efficiency.

The following few examples suggest how knowledge may be integrated as an intermediate step to refine the learned model just before or while arriving at a final model:

- In [57], a neural network is fed sequences with temporal order to iteratively correct the learned model. The approach is claimed to more accurately reproduce small-scale temporal effects.

- In [58], knowledge is integrated through a semantic consistency modifier that improves object detection through a re-optimization step.

- Similarly in [59], a surrogate model is iteratively improved with new observations. This approach is illustrated in Figure 34 and claimed improvements in accuracy, as well as a computational speedup.

- Another way of leveraging knowledge in the final hypothesis is by passing the NN predicted outputs into an embedded knowledge-based model, that can provide a more *informed* loss as the model trains. This sub-class of techniques is covered in greater depth as an embedded and jointly trained approach in Section 4.5.3.

### 4.4.1 Knowledge Integrated into Final Hypothesis for Reinforcement Learning

In reinforcement learning, the reward function is used to weight actions and trajectories that meet some criteria of "goodness" better than other actions and trajectories. In this sense, the reward can be viewed as both the *initial* hypothesis or the *final* hypothesis. Nevertheless, the typical use of the reward function is to reward (or penalize) outcomes based on a designer's initial notion of good behavior of the reinforcement learning system.

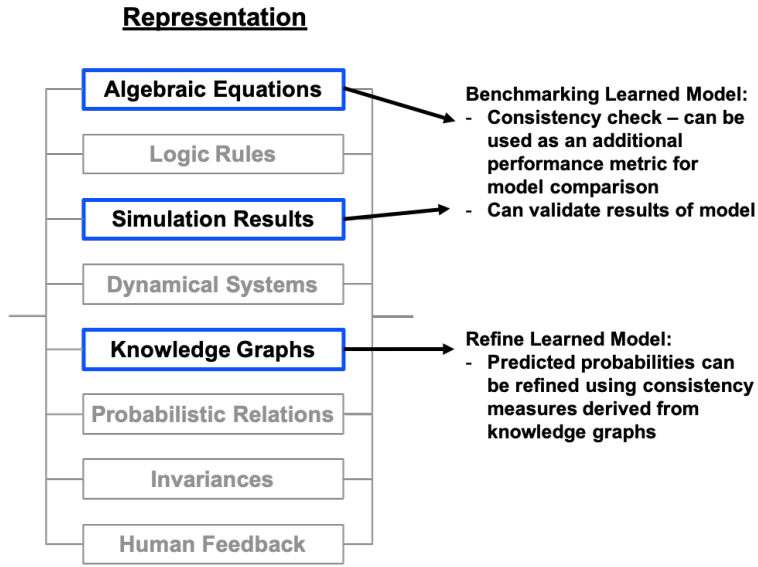

*Figure 33.* Forms of Knowledge Shown to Integrate via Final Hypothesis [1]

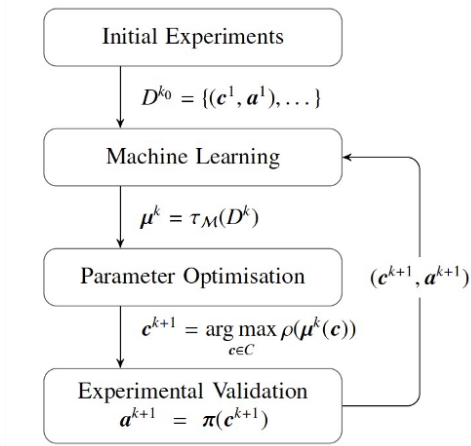

*Figure 34.* Example from Literature that Illustrate Knowledge Integration via Final Hypothesis
This diagram illustrates a training approach from [59] that iteratively improves a surrogate model with new observations.

In some cases, the reward can be used to explicitly inform the reinforcement learning process as a final hypothesis, operating as a consistency check on the results of the learning process. For example, in [60], the authors study branes from string theory. In particular, they wish to find *vacuua*, stable states that can be used for making predictions in physics. The space of plausible vacuua is at least $10^{500}$, and therefore they use a deep RL agent to explore this large space. Not all vacuua are consistent with known physics. The authors use the consistency of each vacuua the system discovers to inform the reinforcement learning process via the reward function. Thus, vacuua that are consistent with known and accepted conditions in physics receive higher reward than those that are inconsistent.

Some selected results from [60] are presented in Fig. 35. First, the authors observed a $\mathcal{O}(200)$ increase in the number of feasible solutions discovered relative to an uninformed method. An additional, qualitative result, is that the consistency checks used in the reward function resulted in a strategy that is similar to how humans think about branes and vacuua. They use so-called "filler" branes to ensure some degree of consistency with known physics, and then test the results of adding other branes. The right-hand side of Fig. 35 displays the results of the RL system learning to use filler branes. These results show that integrating knowledge into the final hypothesis via the reward function is a promising and viable path for RL applications.

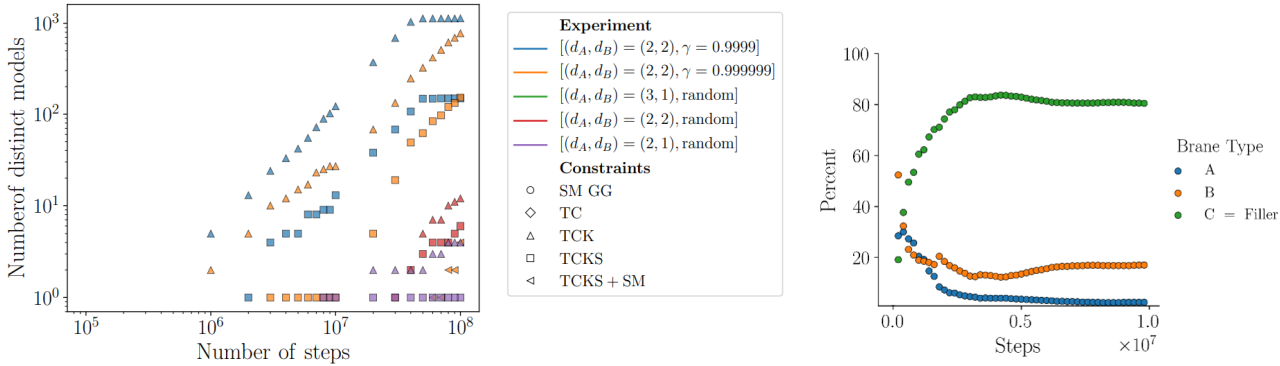

**Figure 35.** *RL Example with Knowledge Integrated via Final Hypothesis*
LEFT: The knowledge-integrated RL approach (orange and blue markers) outperforms naive approaches (green, red, and purple markers) by finding $\mathcal{O}(200)$ more solutions. RIGHT: The knowledge-integrated RL approach learns to use "filler" branes to find a larger set of feasible solutions. [60]

## 4.5 EMBEDDED AND JOINTLY TRAINED NEURAL NETWORKS

A special variant considered in scope for this study, even though the knowledge is not explicitly and tightly integrated, is neural networks (NN) that are embedded and jointly trained. This variant is not explicitly constrained or informed by architecture or topology, rather the NNs are informed by their surrounding context. We found three major categories of embedded and jointly trained neural networks that are described in this section.

### 4.5.1 Neural Network Surrogates

A neural network surrogate is an embedded and jointly trained NN that can take the place of a computationally expensive knowledge-based model, or that can serve to supplement a knowledge-based model in order to represent known, but un-characterized phenomena in the system being described. This strategy exploits the fundamental notion that an NN effectively serves as a universal function approximator, as well as the fact that modern NN architectures are designed in terms of highly efficient, parallelizable array operations. In this way, an NN surrogate can be expected to be both sufficiently expressive as to encapsulate complex and realistic processes in the modeled system, and to be efficient so as to eliminate significant computational bottlenecks from the system.

### 4.5.2 Neural Networks with Surrounding Mathematical Structure

Another flavor of embedded and jointly trained NNs is where the NN is architected to learn unknown terms in an otherwise known mathematical structure, such as a differential equation. In the latter context, the NN can learn to embody specific, unknown mechanisms in a system that is otherwise well characterized by the surrounding informed structure. There can be many benefits to embedding an NN in this way. Generally, provable properties about the encompassing mathematical structure can provide valuable insights into–and even bounds on–the behavior of the NN. Additionally, tools of analysis and methods for working in this context (e.g., numerical methods for solving differential equations) can lead to specialized and principled approaches for training the NN.

"Universal differential equations" (UDEs) is a methodology for incorporating neural networks (or other machine-learnable models) into a differential equation, so that prior structural (e.g., scientific) knowledge can be combined with flexible, data-driven approaches for modeling systems [61]. In a UDE, the neural network is embedded in a differential equation (or a system of equations) alongside known, analytical terms that describe the system's behaviors or dynamics. Thus, in a UDE, the NN is not responsible for learning the governing dynamics of the systems in their entirety, rather it is tasked with modeling the dynamics associated with a subset of the mechanisms at play in the system. In this way, the governing dynamics of the unknown mechanisms or terms can be learned from data, while prior scientific knowledge (e.g., conservation laws, incompressibility constraints, transport models) of the system are modeled explicitly. The embedded NN is trained by solving the differential equations and comparing the result to recorded data. Note that the adjoint sensitivity method enables backpropagation through standard ODE solvers [62], which can be treated as black boxes. Thus the full array of numerical methods for solving differential equations can be leveraged to train the embedded NN.

For example, the UDE methodology for embedding NNs has been demonstrated to enable the recovery of governing equations using less data, to utilize arbitrary conservation laws as prior knowledge in a model, to discover and recover the exact mechanisms that govern a realistic biological

PDE, to enable an adaptive method for solving a 100-dimensional nonlinear PDE, and to automate the modeling of a fast and physically-realistic surrogate of a climate simulation [61].

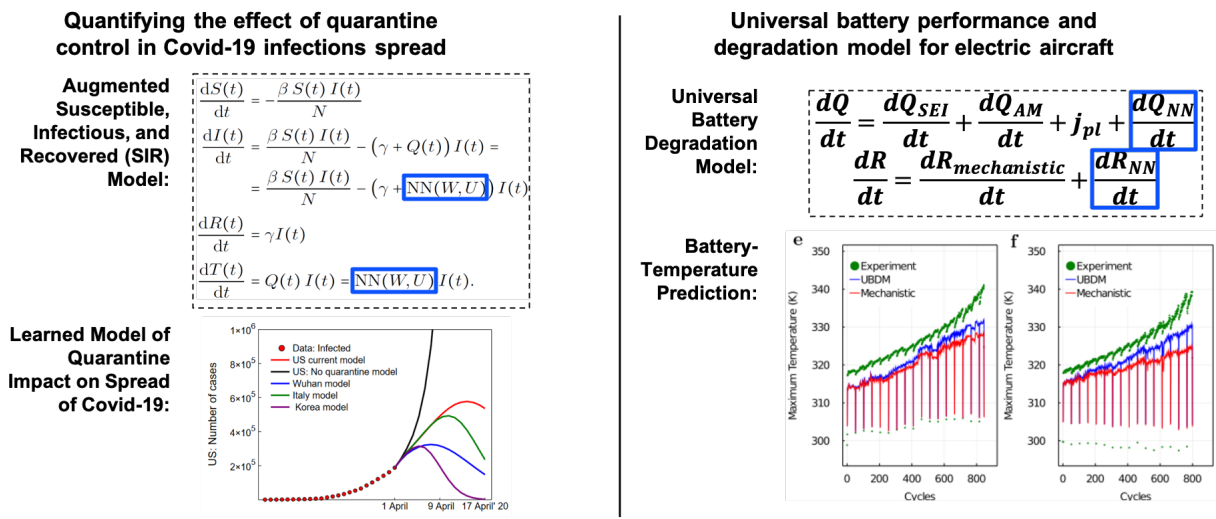

**Figure 36.** *Applications of Universal Differential Equations*
LEFT: A Susceptible, Infectious, and Recovered (SIR) model is augmented with neural network terms to produce a learned model of the quarantine impact on the spread of Covid-19 [63]. RIGHT: A model for battery degradation in electric aircraft combines mechanistic and learned terms to describe the system's dynamics [64].

Although the UDE methodology was only proposed in 2020, it has already made an impact on real-world applications. In the field of infectious disease modeling, a UDE was developed to augment a Susceptible, Infectious, and Recovered (SIR) model to create a learned model of the impact that quarantine policies have on the spread of Covid-19 [63] (Figure 36). A "universal battery degradation model" was used to model the battery performance and degradation for electric aircraft [64]. As depicted in Figure 36, an NN is incorporated with an electrochemical cell model, and various mechanistic models for battery degradation. This research produced a superior physics-informed and data-driven model that better predicts battery degradation and performance than the purely mechanistic battery models.

UDEs can also enable researchers to discover exact analytical forms of dynamical mechanisms in their systems. To demonstrate this, the famed Lotka-Volterra predator-prey model was used to generate noisy population data, and a UDE was formed to explicitly include the prior knowledge of exponential growth (decay) in the prey (predator) while using an NN to represent the unknown population-interaction mechanisms [61].

As shown in Figure 37, the UDE-embedded NN was trained over a narrow domain of training data, which did not encompass a full period of the cyclic solution. The resulting learned predator-prey interaction model proved capable of yielding high-quality extrapolation of the populations far beyond the regime covered by the training data. Furthermore, the authors were able to use the method of sparse identification of nonlinear dynamical systems (SInDy) [65] on the learned component of the UDE to recover–on a polynomial bases–the exact quadratic relationship that describes the predator-prey interaction dynamics in the Lotka-Volterra system.

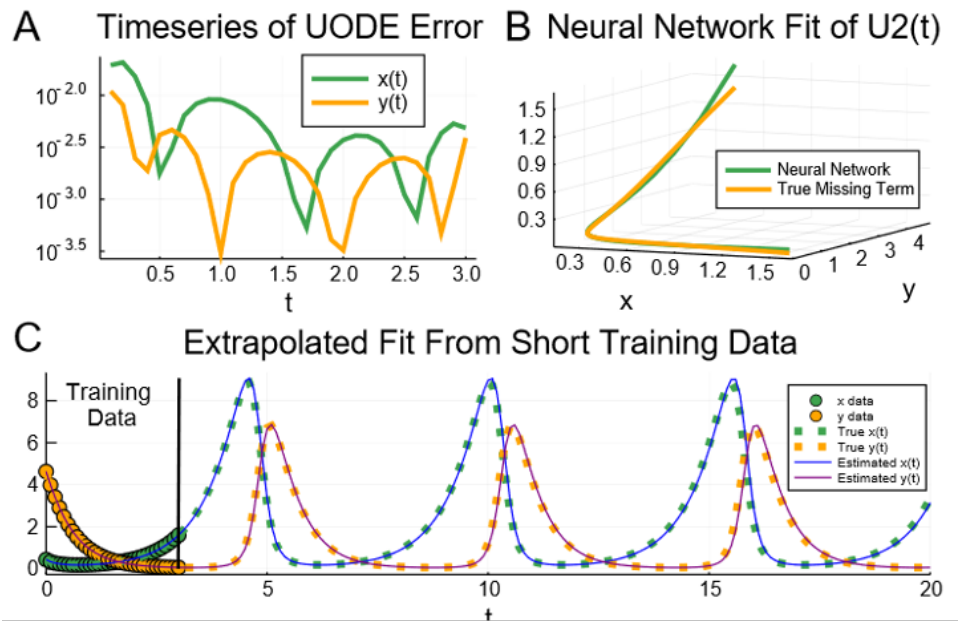

**Figure 37.** *Learning Predator-Prey Dynamics with a Universal Differential Equation*
*A universal differential equation with learned terms for the predator-prey interactions of a Lotka-Volterra system is able to recover, to high fidelity, the quadratic interaction dynamics from only a limited regime of training data.*

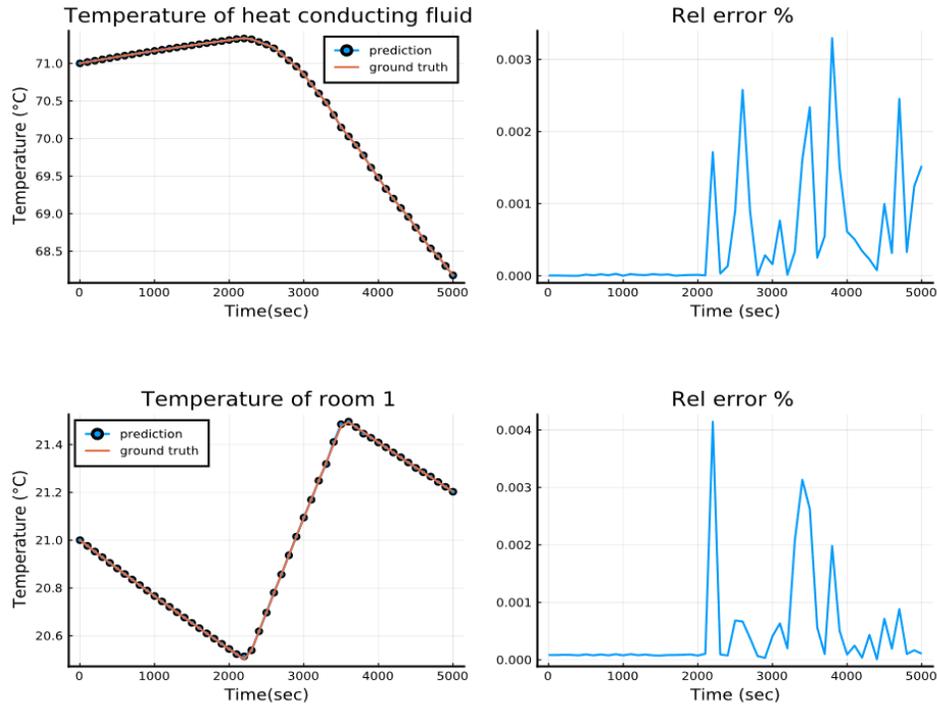

**Figure 38.** *Replacing an Expensive Mechanistic Model with Efficient Surrogate*
*A continuous-time echo state network surrogate of an HVAC system reliably and efficiently models a heating system interacting with ten rooms.*

A learned model embedded in a UDE can also be tailored to the challenging dynamical features of its encompassing system, such as the manifestation of processes on highly disparate time-scales, which is often referred to as "stiffness" in a system. For example, Robertson's chemical reaction equation features stiffness in the form of a rapidly-transforming transient chemical species amidst a slowly-occurring, stable chemical reaction. Models like physics-informed neural networks [66] and recurrent neural networks struggle to learn to be reliable surrogates for such stiff non-linear systems. Thus the embedded model ought to be further informed via its design to try to accommodate these sophisticated dynamics. Continuous-time echo state networks (CTESN) [?] exemplify such an informed design. They prove to be capable of learning highly stiff dynamics, and thus can serve as fast and reliable surrogates for systems that are typically expensive to simulate. A compelling application of a CTESN demonstrated that the learned model was capable of serving as an accurate learned surrogate of a heating, ventilation, and air conditioning (HVAC) system (Figure 38). The runtime of the surrogate scales linearly with system size, whereas the standard implicit ODE solver scales cubically with system size; thus for realistic settings, the learned surrogate accelerates the HVAC simulation by nearly 100x.

### 4.5.3 Consequence Modeling

Consequence modeling is a phrase that we have coined to represent a special case where knowledge is integrated via the final hypothesis of the ML system[1]. Note that this discussion could just as well have been covered in Section 4.4. A consequence model maps the NN's output (e.g., the control parameters for aiming a trebuchet) to a meaningful, measurable quantity of interest (e.g., where a launched projectile ultimately lands, given that the trebuchet was fired using said control parameters). The consequence model (e.g., differential equations describing the kinematics of a trebuchet) is assumed to be differentiable so that its results can be incorporated into the objective function that is used to train the NN; thus the training gradients become informed by the expert knowledge that is instilled via the consequence model. This is in stark contrast to standard approaches to providing feedback to an NN via its final hypothesis; typically, this entails designing a loss function only to have nice mathematical properties for solving a general optimization problem (e.g., an L2 loss measuring the discrepancy between the NN's predicted control parameters against known control parameters). Consequence modeling instead uses expert knowledge to formulate a meaningful and interpretable loss function, whose landscape captures the interplay of quantities of interest–including the NN's various outputs–in a principled way.

Incorporating a differentiable simulator down-stream of a NN is an elegant and powerful way to inform a learned model. Consider, for example, an NN that is responsible for aiming a trebuchet: given the target's position and the wind conditions, the NN must learn to deduce the control parameters (e.g., the counter weight magnitude and angle of release) that will lead to a precise hit. A traditional approach to training a surrogate to this inverse physics problem would be to gather or simulate a large collection of data–control parameter, wind, and target tuples–and then train the NN in a supervised fashion using this training set.

By contrast, an informed approach to this problem was demonstrated in [67] wherein the neural network was placed upstream a differentiable physics model of the trebuchet. In this approach, which is depicted in Figure 39, the neural network is trained by predicting control parameters, which then are used to "fire" the trebuchet. The neural network is thus trained not against a naïve L2 loss between predicted and desired control parameters, but instead against the measured displacement between the desired target and the actual strike point. In this way, the gradients that propagate back to the network are informed by the trebuchet model that is used to measure the meaningful "consequence" of the NN's prediction. As such, the NN is optimized based on the known interplay between disparate quantities (e.g., counter weight and launch angle) that are naturally and equitably combined in forming the consequence-based loss. Not only does this process quickly train a reliable and lightweight surrogate, but it also provides a clear path forward toward improving the system fidelity: rely on scientific expertise to develop increasingly detailed differentiable simulators.

---

[1] Term aptly coined by Jason Nezvadovitz, MIT Lincoln Laboratory, Group 76

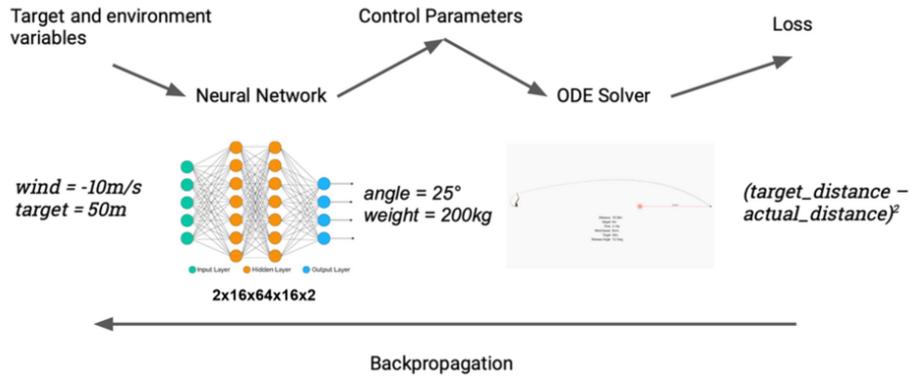

**Figure 39.** *Training a Model to Aim a Trebuchet Using Consequence Modeling*
*A neural network is provided with the wind speed and desired target distance, and is tasked with aiming the trebuchet to hit the target. The model's outputs are verified by simulating the resulting shot with a differentiable physics simulation of the trebuchet.*

Aspects of "consequence" modeling have found early success in realistic control problems. In these applications the consequence can manifest in diverse ways, such as a measured departure from an ideal trajectory [68], or as a violation from a control Hamiltonian [69]. In these cases the objective function can be used to encode directly the constraints of the optimal control problem and thus provide the learned components of system physics-informed feedback to its evolving policy.

This page intentionally left blank.

# 5. PERFORMANCE GAINS VS. VARIANTS OF KNOWLEDGE INFORMED AI

As previously noted (Section 2), knowledge-integrated informed AI techniques have been shown to lead to significant improvements in performance when measured by a variety of metrics that matter for technology problems of national importance (accuracy, enabling new discoveries through broadened applicability, improved safety and reliability, efficient use of resources). Looking closely at which of the forms of knowledge representation and which integration paths lead to these gains in performance suggests that there is a tradespace across informed techniques that should be considered in order to develop the best performing approach for any given problem.

In Table 5, we gather quantified and qualified performance gains reported for each example described in the previous section (and more) into one large table divided by integration path and form of knowledge representation. Comparing reported performance across the rows and columns suggests (1) some gains are achieved regardless of the selected knowledge representation and/or integration path (with differences in magnitude), and (2) some gains are more prevalent for particular knowledge representations and/or integration paths. The following are some observations drawn from this table:

- All integration paths and knowledge forms have demonstrated accuracy or score improvement. The magnitude of these gains suggests that stronger improvements are achieved when knowledge is integrated via the learning algorithm, then hypothesis set, then training data.

- Reporting of improvements in accuracy sometimes coincides with better generalizability beyond training distribution.

- In terms of resource efficiencies:
  - Data efficiency is another improvement that is common across integration paths and knowledge representations, although not always reported on.
  - Claims of improvements in computational efficiency often coincides with data efficiency. Also note that knowledge integration can instead increase compute requirements. This has at least been reported in two cases where knowledge is integrated into the learning algorithm [55] [70].
  - Effect on network size (i.e., network simplicity) is not commonly measured. Few reports suggest effect on size can be better (fewer parameters, smaller network) or worse (more parameters, larger network). Furthermore, we found that this effect is only reported when the knowledge form is invariances.

- Claims of interpretability are made mostly when the knowledge representation has some symbolic reasoning power. An exception is where the network architecture itself is based on knowledge.

- Informed RL examples mostly use symbolic representations of knowledge (logic rules, knowledge graphs, and human feedback) and often claim improvements in interpretability (especially if integration path is via hypothesis set).

- There is relatively less research activity for knowledge integration into training data and hypothesis set. This may shift with greater awareness of informed techniques.

There are several factors to note as we draw conclusions from this table:

1. Researchers more often report on performance gains, not disadvantages.

2. Reporting on some metrics is standard (accuracy, score improvements, reductions in test error), however reporting on other metrics (generalizability, data efficiency) is entirely up to the authors.

3. Several observed reporting inconsistencies:

    - Improvements are not always reported using the same metrics—overall accuracy, AUROC, RMSE, test error, mAP, IoU, etc.
    - Certain metrics are rarely or never quantified (generalizability, interpretability, reliability).
    - Choice of baseline model is sometimes questionable or missing entirely.

4. Publications generally document unique solutions for different problems. Therefore, while comparing performance gains across solutions suggests an interesting tradespace, theoretical and empircal analysis across various techniques for select problems, or classes of problems, would be more conclusive.

# TABLE 5

**Performance Gains by Knowledge Representation and Integration Path**

1. Accuracy (%a), Test error (%e) / Score (xS)
2. Generalizable (G) / Interpretable (I)
3. Resource Efficiency (%<D data, %<L labels, xC computational speed-up, fraction N of network size)

White Box Highlight = RL Examples
[+]Only qualified (or quantified with unique metric)
1st sh = First super-human performance

| | Training Data | | | Hypothesis Set | | | Learning Algorithm | | | Final Hypothesis | | |
|---|---|---|---|---|---|---|---|---|---|---|---|---|
| | 1 | 2 | 3 | 1 | 2 | 3 | 1 | 2 | 3 | 1 | 2 | 3 |
| Algebraic Equations | | | | 2-16%a, 30-60%e | | 440xC, 10xC | -4-43%e | | 0%<L | %a[+] | | |
| Logic Rules | | | | xS[+], 8xS | I[+], I[+], I[+] | 57%<D, <D[+],xC[+], 100xC | 20%a, 8xS | I[+] | 0%L, 1%L | | | |
| Simulation Results | 1-7%a, 3-5xS | G[+] | | | | 40%<D, 100xC | %a[+], %a[+] | G>, G[+] | <D[+], >C | %a[+] | | xC[+] |
| Dynamical Systems | | | | %a[+], same, -99%e | G[+] | <D[+], xC[+], <D[+], 15kxC, 2xC | -30%e, %a[+], %a[+], 1st sh | | <D[+], 96%<D, 4xC | | | |
| Knowledge Graphs | .6-1.7%a, 2.4%a | I[+] | | .6-1.7%a, 2.4%a, 8xS | I[+], I[+] | | Discovery, 1.5-6.5%a, 8xS | I[+], G[+], I[+] | | %a[+], %a[+] | G[+], G[+] | |
| Probabilistic Relations | | | | 9-10%a,-50%e, Efficacy | G[+], G[+] | <D¡D¿ | | | | | | |
| Invariances | %a[+] | G[+] | <D[+], xC[+] | -6%e, -30%e, -3.3%e | G[+], I[+] | 1/3rdxN, 1.5xN, xC[+] | | I[+] | xC[+] | | | |
| Human Feedback | %a[+], %a[+], 1-4.5xS, 1st sh | | <D[+], xC[+] | %a[+], 1.5-3xS, 1-4.5xS, 1st sh | I[+] | <D[+], xC[+] | 1st sh, 1st sh, 1.4-6.5xS | Trust[+] | xC[+], <D[+] | | | |

## TABLE 6

**References Associated with Gains Included in Performance Gains Table**

1. **Accuracy (%a), Test error (%e) / Score (xS)**
2. **Generalizable (G)** / **Interpretable (I)**
3. **Resource Efficiency (%<D data, %<L labels, xC computational speed-up, fraction N of network size)**

White Box Highlight = RL Examples
[+]Only qualified (or quantified with unique metric)
1st sh = First super-human performance

| | Training Data | | | Hypothesis Set | | | Learning Algorithm | | | Final Hypothesis | | |
|---|---|---|---|---|---|---|---|---|---|---|---|---|
| | 1 | 2 | 3 | 1 | 2 | 3 | 1 | 2 | 3 | 1 | 2 | 3 |
| Algebraic Equations | | | | [44] [71] | | [71] [72] | [49] | | [73] | [57] | | |
| Logic Rules | | | | [74], [75] | [32], [74], [75] | [76], [32], [32], [32], [48] | [77], [75] | [75] | [73], [77] | | | |
| Simulation Results | [35], [39] | [35] | | | | [78], [48] | [79], [55] | [79], [55] | [55], [55] | [59] | | [59] |
| Dynamical Systems | | | | [51], [80], [80] | [47] | [47], [47], [51], [80], [80] | [50], [51], [81], [4] | | [51], [82], [82] | | | |
| Knowledge Graphs | [34], [43] | [43] | | . [34], [43], [75] | [43], [75] | | [53], [54], [75] | [53], [54], [75] | | [83], [58] | [83], [58] | |
| Probabilistic Relations | | | | [45], [45], [46] | [45], [46] | [51] | | | | | | |
| Invariances | [38] | [38] | [38], [38] | [41], [40], [42] | [40], [84] | [41], [42], [84] | | [84] | [84] | | | |
| Human Feedback | [85], [36], [86] | | [86], [86] | [85], [87], [86], [86] | [87] | [86], [86] | [4], [88], [89] | [56] | [88], [70] | | | |

# 6. CHALLENGES AND UNTAPPED RESEARCH OPPORTUNITIES

For a lengthy set of challenges and directions for knowledge-integrated informed AI approaches, we refer to the last section in Laura von Rueden et. al.'s survey paper [1]. Here we offer a brief collection of promising research directions and application areas we found to be untapped and worthwhile for advancing technology for national security. Each item here stands as a gap in foundational research, or an applied technology opportunity.

- The tradespace across knowledge-integrated informed AI techniques (employing different knowledge representations and integration paths) has been barely explored beyond drawing comparisons across dispersed research activities. One example paper from more than two decades ago discussed the difference between different integration paths and suggested that "expanding the effective training set size ... is mathematically equivalent to incorporating the prior knowledge as a regularizer" [38]. Exploring the tradespace theoretically and empirically, considering each form of knowledge representation, their plausible integration paths, and how much importance to give to the knowledge vs. data will help lead to the development of techniques that are most optimal for any given problem. Others have also suggested this as an important research direction noting that "as the field of AI moves towards agreement on the need for combining the strengths of neural and symbolic AI, it should turn next to the question: what is the best representation for neurosymbolic AI?" [10] This sentiment can, of course, be extended beyond just neurosymoblic AI.

- Performance *degradation* caused by knowledge integration is largely unreported. For example, it is plausible that knowledge integration can be too constraining, or perhaps even misleading, precluding learning. Alternatively, knowledge integration could be a computational bottleneck that slows down training and/or learning. These kinds of issues are yet to be explored and understood.

- Knowledge representation in deep learning may not scale well, therefore there is a need to understand the limits and the reasons for those limits. In other words, "as we expand and exercise the symbolic part and address more challenging reasoning tasks, things might become more challenging... For example, among the biggest successes of symbolic AI are systems used in medicine, such as those that diagnose a patient based on their symptoms. These have massive knowledge bases and sophisticated inference engines. The current neurosymbolic [informed] AI isn't tackling problems anywhere nearly so big." [90] This observation too can be extended beyond just neurosymbolic AI.

- It's plausible that the integration of knowledge that constrains or guides learning would result in solutions that are more robust to adversarial attack. Analysis that specifically explores sensitivity to such counter-AI techniques with and without knowledge integration may show strong ability to address concerns of robustness in today's AI with knowledge-integrated informed AI.

- Verification and validation of engineered systems is an exponential problem in general, and becomes more so for systems that employ AI. It's plausible to suggest that explicitly informed techniques are easier to verify/validate. Developing proof of this hypothesis along with specific methodology for testing informed systems can help make strides toward safer AI.

- The recent dramatic successes of transformer-based, self-supervised models in the domains of both computer vision (CV) and natural language processing (NLP) may serve as a blueprint for developing foundational models for scientific domains that are informed by rich simulation frameworks. It is an open question as to whether a traditional, attention-based transformer architecture will be as impactful in problem spaces where informed AI techniques are especially promising, or if such architectures and learning paradigms can be augmented to explicitly incorporate expert knowledge.

- Updating of knowledge that is explicitly integrated could enable AI systems that are more adaptable via continuous learning and continuous integration. This research direction would be valuable for learn-in-the-field (edge-AI) applications.

- There's a dependency on programming languages and packages that allow for knowledge integration. Advancements in the state of the art of knowledge-integrated informed AI therefore depend on continued efforts to develop languages and packages that allow all kinds of knowledge integration. This issue is discussed in more detail below.

**A Language Problem** Among the many research projects examined here, the vast majority of them are implemented using one of a handful of Python-based libraries for deep learning. Over the past decade, Python has emerged as the language of choice among data scientists and machine learning researchers; its popularity has earned it a critical mass of users such that a rich, unparalleled ecosystem of powerful and polished tools, libraries, tutorials, and reference implementations has manifested, with the likes of Microsoft and Facebook having invested heavily into it. It is this Python-centric ecosystem that has fostered the recent deep learning revolution in domains like computer vision and natural language processing. But is this sustainable? Is Python a language that can foster the expansion of novel applications of deep learning into other scientific fields?

The fact of the matter is that all scientific computing Python libraries, in one way or another, depend on a faster, "lower-level" language–like C–to handle all of the fast mathematical computations. Thus far, deep learning models for computer vision and natural language processing applications have not been so hindered by this "two language" paradigm; dense, convolutional, and recurrent neural networks can be framed in terms of vectorizable, linear algebra routines, which can easily be dispatched to a lower-level language. Indeed these are the corner stones of libraries like PyTorch and TensorFlow. However, as the disciplines of scientific computing and machine learning meet and begin to cross-pollinate, there will be a need to support fast computation and automatic differentiation involving adaptive algorithms, non-standard algebras (e.g., quaternion algebra), and the ability to interface with other programming paradigms, such as probabilistic programming languages. In short: knowledge-integrated informed AI will likely lead us away from the land of multiplying massive matrices of real numbers.

Python, in its present form, can never address such gaps itself. It can only be used as an interface with other languages that can then attempt to solve these problems. This is why Jeremy Howard, creator of the fast.ai deep learning library (which is Python-based, of course) asserted:

"Python is not the future of machine learning. It can't be. It's so nicely hackable, but it is so frustrating to work with a language where you can't do anything fast enough unless you call out to some external CUDA or C code [...] It's fine for a lot of things, but not really in the deep learning or machine learning world. I really hope that Julia is successful [...] Most importantly, it's Julia all the way down, so you can write your GPU kernel in Julia." [91]

Google's JAX [92] is a Python library that is growing in popularity as a route toward efficient, differentiable programming; it provides an XLA-based (Accelerated Linear Algebra) just-in-time compiler that can operate on a subset of Python and NumPy to accelerate computations that would otherwise fall outside of the "vectorizable" suite of code patterns. JAX makes it simple to write fast, compose-able NumPy-based functions, and it provides a rich automatic differentiation framework for accessing various high-order derivatives, Jacobian-vector products, etc., of these functions. As such, it is being used in some pioneering works in the knowledge-integrated informed AI space, such as the development of a differentiable molecular dynamics simulator [93] and a differentiable quantum chemistry simulator (Li, 2021). However, JAX is ultimately limited by its roots in the Python language; simple control-flow logic and in-place operations on data structures are not supported by JAX, which limits its applicability to many domains of scientific computing and thus knowledge-integrated informed AI.

Beyond the world of Python, Julia is a programming language that is designed to be just-in-time compiled such that performant code can be written natively in Julia, while still being accessible and readable like Python. Julia is likely the language that boasts the most comprehensive automatic differentiation capabilities; for instance, the Zygote automatic differentiation library operates on Julia's abstract syntax tree itself to enable differentiable programming [67]. In this way, Julia is similar to but exceeds JAX in a fundamental way: it is unimpeded by the two (or N) language problem. For instance, one can easily write highly performant Julia code that operates on exotic data types using non-standard algebras and then use this code in a larger machine learning framework. Additionally, Julia is designed around a multiple dispatch system [94], which fosters rich code reuse such that features like automatic differentiation, GPU dispatch, and fast linear algebra routines can easily be extended to one's novel mathematical framework. In terms of design, Julia is the most natural language to do innovative research in the realm of knowledge-integrated informed AI. But the cost of leaving the rich ecosystem of Python-based machine learning and data science resources is difficult to overstate. It has both practical and far-reaching implications ranging from Julia's relative lack of infrastructure and convenient tooling, to making one's research inaccessible to others in the field, to Julia being unknown or untested by the software security standards that are required for deploying solutions in secure environments.

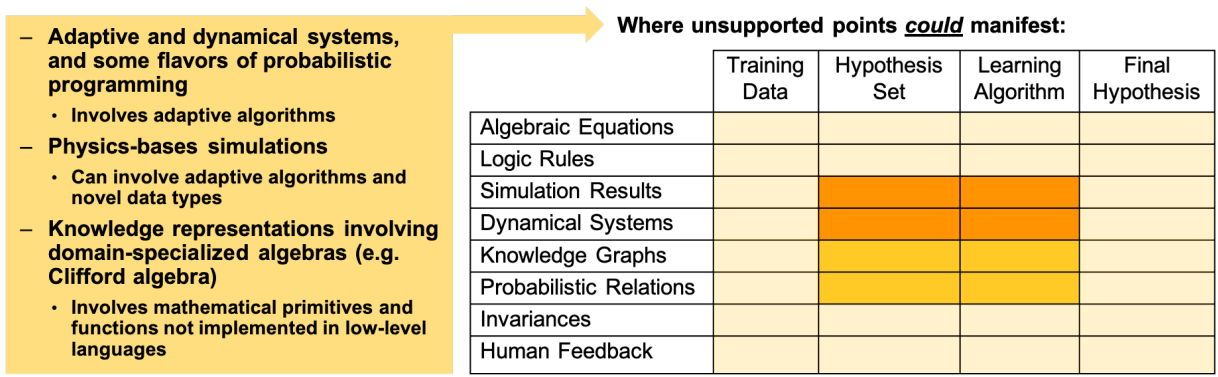

**Figure 40.** *A Programming Language Problem*
*A heat map that indicates the approaches to informed AI that are most likely to be obstructed by the "two language problem." To tightly incorporate simulations and models of dynamical systems into one's ML framework is often a substantial challenge in Python, whereas Julia is highly capable toward this end.*

# 7. SUMMARY

AI that leverages both data and knowledge is continuously gaining interest as it delivers noteworthy and game-changing performance. Within the national security domain, where there is ample scientific and domain-specific knowledge that stands ready to be leveraged, and where purely data-driven AI can lead to serious unwanted consequences, this emerging class of AI techniques should be deliberately pursued.

In this report, we discuss the motivations behind what we refer to as "knowledge-integrated informed AI", along with our findings from a thorough exploration of approaches that fall into this paradigm of AI. Through illuminating examples of knowledge integrated into both deep learning and reinforcement learning pipelines, we shed light on specific kinds of performance gains that knowledge-integrated informed AI has been shown to provide. We also discuss an apparent trade space across variants, along with observed and prominent issues that suggest worthwhile future research directions. Most importantly, we discuss how performance gains can benefit problems of national security, and we lay out immediate and longer term steps we can take to appropriately adapt and apply knowledge-integrated informed AI techniques.

This page intentionally left blank.